\documentclass{article} 
\usepackage{iclr2027_conference,times}

\usepackage{amsmath,amsfonts,bm}

\def\eqref#1{equation~\ref{#1}}

\def\1{\bm{1}}

\DeclareMathAlphabet{\mathsfit}{\encodingdefault}{\sfdefault}{m}{sl}
\SetMathAlphabet{\mathsfit}{bold}{\encodingdefault}{\sfdefault}{bx}{n}

\usepackage{hyperref}
\usepackage{url}

\usepackage{booktabs}
\usepackage{multirow}
\usepackage[table]{xcolor}
\usepackage{adjustbox}
\usepackage{arydshln}
\usepackage{amssymb}

\usepackage{algorithm}
\usepackage{algorithmicx}
\usepackage{algpseudocode}

\usepackage{xspace}
\usepackage{wrapfig}
\newcommand{\method}{CrystalAR\xspace}
\definecolor{symred}{RGB}{255,242,242}
\usepackage{makecell}
\definecolor{lightgreen}{RGB}{220, 245, 220}

\title{CrystalAR: Autoregressive Modeling for Symmetric Crystal Generation}

\author{
{\bfseries
Ziqiao Meng$^{1,2}$\thanks{Equal contribution.},
Rongsheng Chen$^{3}$\footnotemark[1],
Zhi Zheng$^{2}$\footnotemark[1],
Nikita Kazeev$^{1}$,
Zhenkun Wang$^{3}$\thanks{
Corresponding authors:
\texttt{wangzk3@sustech.edu.cn},
\texttt{dcsleews@nus.edu.sg}.
}}\\
{\bfseries
Andrey Ustyuzhanin$^{1}$,
Wee Sun Lee$^{2}$\footnotemark[2]}\\
$^{1}$Institute for Functional Intelligent Materials (IFIM),
National University of Singapore\\
$^{2}$School of Computing, National University of Singapore\\
$^{3}$Southern University of Science and Technology
}

\iclrfinalcopy 
\begin{document}

\maketitle
\lhead{Preprint}

\begin{abstract}

Deep generative modeling has achieved remarkable progress in crystal generation. Given the rich symmetries inherent in crystalline structures, recent work has sought to explicitly incorporate space-group symmetry into generative models, enabling more effective exploration of novel symmetry templates. Yet introducing symmetry templates increases the combinatorial complexity of stoichiometric relationships between atomic species and crystallographic sites, while creating a fundamental tension between flexible exploration of discrete crystallographic templates and faithful realization of these templates as continuous three-dimensional structures. To address these challenges, we propose \textbf{\method}, a unified autoregressive framework that models Wyckoff symmetry templates and complete crystal structures within a single generative sequence. Specifically, \method reformulates continuous coordinate generation as a symmetry-conditioned hierarchy of discrete spatial localization decisions, autoregressively refining the positions of all symmetry-inequivalent sites across resolution levels while jointly modeling their spatial configuration within each level. This discrete hierarchical formulation further enables constraint-aware search over both crystallographic templates and geometric realizations. As a result, \method retains strong symmetry-template generation capability while achieving competitive de novo generation performance and template-free crystal structure prediction (CSP) accuracy. These results suggest that symmetry-template generation can be retained alongside broadly competitive de novo generation performance and competitive CSP without reliance on symmetry templates retrieved from prior data, providing a promising alternative paradigm for symmetry-preserving crystal generation. \footnote{Project Homepage:
\href{https://crssbl.github.io/Project-CrystalAR/\#results}
{\nolinkurl{https://crssbl.github.io/Project-CrystalAR/\#results}}}

\end{abstract}

\section{Introduction}

Discovering new crystal structures is fundamental to materials innovation. However, the vast combinatorial space spanning chemical compositions and atomic configurations makes exhaustive exploration of stable and novel crystals prohibitively expensive~\citep{curtarolo2013high,pickard2011ab}, motivating computational approaches that can efficiently navigate this enormous search space~\citep{MP,AFLOW}. In recent years, deep generative modeling for crystal generation~\citep{MatterGen,DiffCSP,FlowMM,CDVAE} has attracted increasing attention due to its strong potential to efficiently explore this vast design space and accelerate the discovery of novel materials. By learning the underlying distribution of experimentally observed or computationally characterized crystal structures, generative models~\citep{Diffusion,Flow-matching,sto-interpolant} have achieved initial success in directly proposing diverse and physically plausible candidates. 

Crystalline materials are fundamentally characterized by rich crystallographic symmetries~\citep{nye1985physical,aroyo2006bilbao}, which organize atomic configurations into space groups and strongly constrain their structural and physical properties~\citep{ITC-table,bradlyn2017topological}. Accordingly, recent crystal generative models incorporate symmetry either through explicit crystallographic constraints in specialized symmetry-template-constrained models~\citep{DiffCSP++,SGFM}, or through softer space-group conditioning in more general-purpose generative frameworks, such as MatterGen~\citep{MatterGen}. However, symmetry-constrained generation alone does not necessarily imply meaningful exploration of crystallographic structure space. Recent studies have highlighted that conventional structural novelty can coexist with substantial reuse of previously observed symmetry templates, motivating template-level novelty as a complementary measure of structural exploration~\citep{negishi2026substitution,WyFormer,SymmCD}. This observation has motivated a growing line of work that explicitly generates Wyckoff configurations, including WyFormer~\citep{WyFormer}, WyckoffDiff~\citep{WyckoffDiff}, SymmCD~\citep{SymmCD}, SGEquiDiff~\citep{SGEquiDiff}, and CrystalFormer~\citep{CrystalFormer}. By directly modeling the discrete Wyckoff-template space, these approaches enable explicit exploration of crystallographic configurations beyond selecting or conditioning on predefined symmetry patterns. We refer to this class of methods as \textit{symmetry-native models}.

\begin{figure*}[!t]
    \centering
    \includegraphics[width=0.99\textwidth]{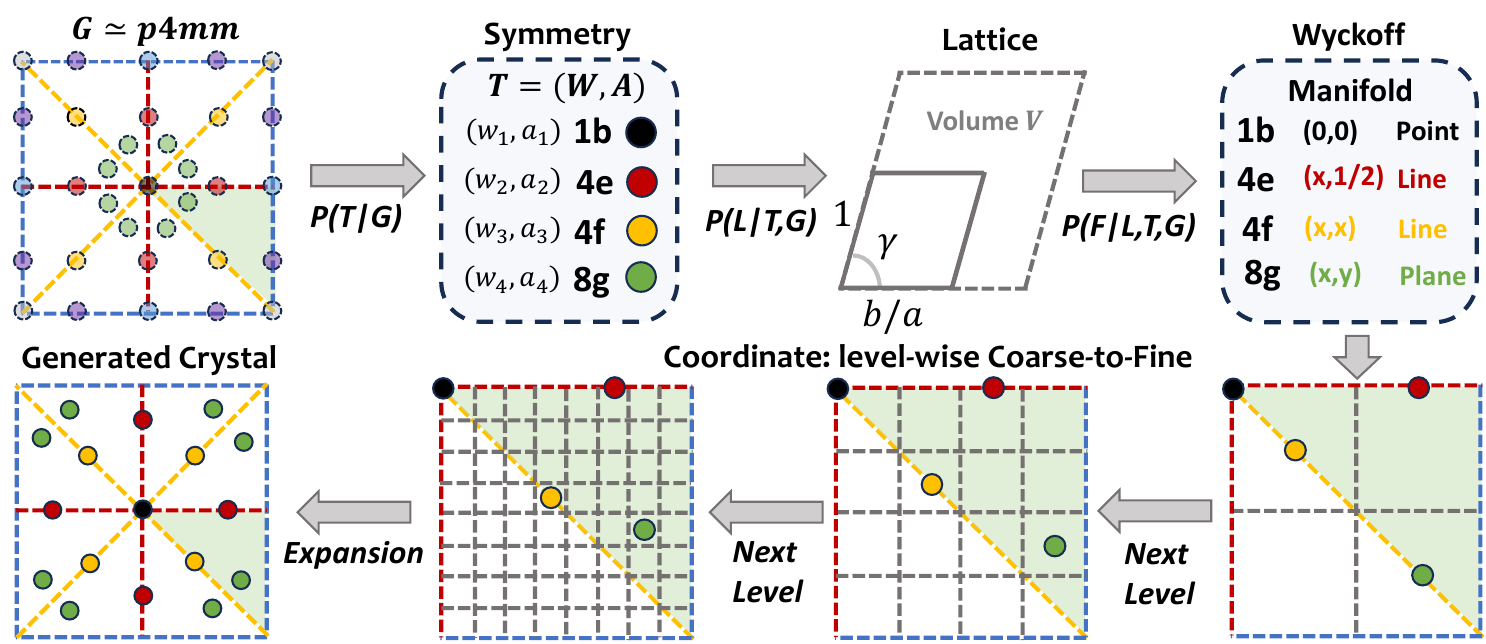}
    \caption{Overview of \method. Given a symmetry group $G$, \method autoregressively generates the element-decorated Wyckoff template $T=(W,A)$, lattice parameters $L$, and free Wyckoff coordinates $F$. Each inequivalent site is restricted to its corresponding Wyckoff manifold. The free coordinates are then generated through $\textit{level-wise coarse-to-fine}$ refinement, and symmetry operations expand them into the complete crystal structure.}
    \label{fig:CrystalAR-Figure01} 
\vspace{-20pt}
\end{figure*}

Explicit template generation, however, introduces new challenges for both de novo generation and crystal structure prediction (CSP). First, the combinatorial Wyckoff-template space must be reconciled with the strict geometric constraints imposed by each template, making it difficult to preserve flexible template exploration while consistently realizing valid and stable crystal geometries. This tension can manifest as degraded de novo generation performance~\citep{CrystalFormer,SymmCD} or necessitate external relaxation procedures for complete geometric realization~\citep{DFT,chgnet}. Second, CSP imposes stricter requirements: the generated template must exactly satisfy the prescribed composition through Wyckoff multiplicities, while the resulting geometry must closely recover the target structure. Consequently, symmetry-native models may either rely on retrieved templates and external lattice samplers to narrow the search space~\citep{SGEquiDiff} or struggle to accurately recover the ground-truth structure while retaining flexible template generation~\citep{WyFormer,CrystalFormer,SymmCD,WyckoffDiff}.

In this work, we propose \method, a unified autoregressive framework that hierarchically generates crystals from discrete symmetry configurations to three-dimensional geometry. Given a space group, \method first generates an element-decorated Wyckoff template and then realizes it through symmetry-conditioned, coarse-to-fine localization of fractional coordinates, jointly resolving all symmetry-inequivalent sites at each resolution level, as illustrated in Figure~\ref{fig:CrystalAR-Figure01}. This formulation naturally supports structured decoding, combining constraint-guided template search with hierarchical exploration of geometric realizations. On LeMat-GenBench~\citep{lemat}, \method achieves strong S.U.N. performance both with and without pre-relaxation, together with competitive S.S.U.N. scores under symmetry-aware evaluation. It also achieves competitive CSP accuracy without relying on Wyckoff-template retrieval. Overall, \method retains strong template-generation capability while achieving broadly competitive de novo generation performance and template-free CSP accuracy, representing a substantial step forward in both the practical performance and generative flexibility of symmetry-native crystal generation. Collectively, these capabilities motivate a paradigm of \emph{autoregressive crystallographic search}, where symmetry templates and their geometric realizations are explored within a single generative process, providing a promising alternative direction for symmetry-preserving crystal generation.

\section{Related Works}


\paragraph{Crystal Generation.} Deep generative modeling for crystal structure generation has attracted considerable attention in recent years. Diffusion-based models~\citep{CDVAE,DiffCSP,MatterGen,ADIT,UniMat,KLDM} have emerged as the dominant paradigm, with numerous extensions tailored to the periodic and three-dimensional nature of crystalline materials. In parallel, flow-matching approaches~\citep{FlowMM,FlowLLM,OMG,CrystalFlow,MCFlow} have been actively explored, offering improved training and sampling efficiency as well as greater flexibility in model design. Motivated by the success of large language models (LLMs)~\citep{GPT-3,GPT-4}, several recent works have further investigated crystal generation through CIF-based language modeling~\citep{UniGenX,NatureLM,CrystalLLM,plaid}, specialized tokenization schemes, and preference-aligned post-training. Despite strong performance on standard generation benchmarks, these generic approaches often capture crystallographic symmetry only implicitly, and their apparent structural novelty can largely arise from elemental substitutions of known structural prototypes rather than the discovery of genuinely novel symmetry templates~\citep{negishi2026substitution}. 

\paragraph{Symmetric Crystal Generation.} Wyckoff positions~\citep{ITC-table} introduce additional discreteness into crystal representations, yielding a heterogeneous space that combines discrete variables, such as elements and Wyckoff positions, with continuous lattice and coordinate variables. Existing methods address this heterogeneity in different ways. DiffCSP++~\citep{DiffCSP++} and SGFM~\citep{SGFM} retrieve or sample Wyckoff templates from training data and focus primarily on template-conditioned geometry generation, whereas WyCryst~\citep{WyCryst}, WyFormer~\citep{WyFormer}, and WyckoffDiff~\citep{WyckoffDiff} mainly generate element-decorated Wyckoff templates and rely on external relaxation or geometry refinement to obtain complete structures. SGEquiDiff~\citep{SGEquiDiff} models both components but learns the discrete template and continuous geometry separately. CrystalFormer~\citep{CrystalFormer} and SymmCD~\citep{SymmCD}, by contrast, jointly model these heterogeneous variables within a single generative process using autoregressive modeling and discrete diffusion, respectively. Our \method takes a different approach by discretizing Wyckoff-constrained fractional coordinates and jointly modeling symmetry templates and crystal geometry in an end-to-end autoregressive framework.


\section{Preliminary}

\paragraph{Crystal Representations.} A periodic crystal structure can be represented by its lattice, atomic species, and atomic coordinates. Let $\mathbf{L}\in\mathbb{R}^{3\times3}$ denote the lattice matrix and $\lbrace(e_i,\mathbf{f}_i)\rbrace_{i=1}^{N}$ the atomic species and fractional coordinates, where $\mathbf{f}_i\in[0,1)^3$ specifies the position of atom $i$ within the periodic unit cell. The corresponding Cartesian coordinate is given by $\mathbf{x}_i=\mathbf{L}\mathbf{f}_i$, with $\mathbf{L}$ determining the physical shape and scale of the cell. Equivalently, the lattice can be described by the three lattice lengths and three interaxial angles, $(a,b,c,\alpha,\beta,\gamma)$. For modeling, we further reparameterize these variables as $\boldsymbol{\ell}=(V,r_b,r_c,\alpha,\beta,\gamma)$, where $V$ is the unit-cell volume, $r_b=b/a$ and $r_c=c/a$ encode the relative lattice lengths, and $\alpha,\beta,\gamma$ denote the lattice angles. This parameterization separates the global lattice scale from its shape while preserving the full six degrees of freedom of the unit cell. Further details are provided in the Appendix~\ref{appendix:lattice}. 

\paragraph{Space-Group Symmetry \& Wyckoff Positions.}
A space group $G$ specifies a collection of affine symmetry operations
$g=(\mathbf{R}_g,\mathbf{t}_g)$ acting on a fractional coordinate as $g(\mathbf{f})
= \mathbf{R}_g\mathbf{f}+\mathbf{t}_g \pmod{1}.$ Starting from a representative site $\mathbf{f}$, these operations generate a set of symmetry-equivalent positions, $\mathrm{Orb}_{G}(\mathbf{f}) = \{g(\mathbf{f}) \mid g\in G\}$. The local symmetry of the representative site is described by the subgroup $\mathrm{Stab}_{G}(\mathbf{f}) = \{g\in G \mid g(\mathbf{f})\equiv\mathbf{f}\pmod{1}\}$, which contains the operations that preserve the site modulo lattice translations. Wyckoff positions group sites with equivalent stabilizer structure and therefore provide a discrete characterization of the symmetry constraints imposed on atomic locations. For a Wyckoff position $w$, its multiplicity $m(w)$ gives the number of sites generated in the corresponding orbit, while its positional degree of freedom $d(w)\in\{0,1,2,3\}$ specifies the number of independent fractional-coordinate parameters required to locate its representative site. Consequently, a crystal can be reduced to $M$ symmetry-inequivalent representatives, $\{(e_i,w_i,\mathbf{f}_i)\}_{i=1}^{M}$, from which all atoms in the unit cell are reconstructed by symmetry expansion. We define $T=(A, W) = \{(e_i,w_i)\}_{i=1}^{M}$ as the \emph{element-decorated Wyckoff template}, separating the discrete crystallographic configuration from its coordinate realization.

\section{Methodology: \method}

\subsection{Autoregressive Modeling for Symmetric Crystal Structures}

\paragraph{Symmetry-Constrained Geometry.}
Crystallographic symmetry restricts crystal geometry at both the lattice and coordinate levels. The space group $G$ determines the admissible lattice family; for example, a tetragonal lattice satisfies $a=b\neq c$ and $\alpha=\beta=\gamma=90^\circ$, while a cubic lattice further requires $a=b=c$. At the atomic level, each Wyckoff position $w\in W$ constrains its representative fractional coordinate $\mathbf{f}$ to a symmetry-defined manifold. A special position may be fixed at $\mathbf{f}=(0,0,0)$, whereas line- and plane-constrained positions retain only one or two free parameters, such as $(x,\frac{1}{2},\frac{1}{2})$ and $(x,y,0)$. Thus, symmetry progressively reduces the feasible geometric space, from the lattice family specified by $G$ to the coordinate manifolds specified by $W$.

These constraints naturally induce a directional dependency among the crystal variables. Given $G$, the model first determines the element-decorated Wyckoff template $T=(A,W)$, which specifies the chemical occupancy and symmetry manifolds of the inequivalent sites. The lattice parameters $\mathbf{L}$ are then generated within the lattice family permitted by $G$, conditioned on $T$ to determine a specific lattice realization. Finally, the fractional coordinates $\mathbf{F}$ are localized within the manifolds defined by $W$. We place $\mathbf{L}$ before $\mathbf{F}$ because the lattice determines the metric geometry---and hence interatomic distances---associated with fractional coordinates. This directional structure motivates the autoregressive factorization
\begin{equation}
p(G,A,W,\mathbf{L},\mathbf{F})
=
p_{\mathrm{emp}}(G)\,
p(T\mid G)\,
p(\mathbf{L}\mid T,G)\,
p(\mathbf{F}\mid \mathbf{L},T,G),
\end{equation}
where $p_{\mathrm{emp}}(G)$ denotes the empirical space-group distribution estimated from the training data. We begin by constructing the discrete \emph{symmetry skeleton} of the crystal through the element-decorated template $T=(A,W)$. Each generated site fixes both a chemical identity and a symmetry orbit, progressively narrowing the remaining combinatorial space. Within each site, we first predict the element $e_i$ and then its Wyckoff position $w_i$, conditioning the symmetry assignment on the current chemical choice. To reduce permutation ambiguity, sites are ordered from higher to lower Wyckoff site symmetry, with special positions preceding less constrained line- and plane-like manifolds. A learned stopping decision determines when the template is complete:
\begin{equation}
\begin{aligned}
p(T\mid G)
=
\prod_{i=1}^{M}
& p(e_i \mid e_{<i},w_{<i},G)\,
p(w_i \mid e_{\leq i},w_{<i},G)\,
p\!\left(
s_i=\mathbb{I}[i=M]
\mid e_{\leq i},w_{\leq i},G
\right).
\end{aligned}
\end{equation}

Once the symmetry skeleton $T$ is fixed, the lattice $\mathbf{L}$ determines the global metric in which this skeleton is geometrically realized. We generate the six lattice variables autoregressively, allowing each parameter to condition on both the symmetry context and previously determined lattice geometry:
\begin{equation}
    p(\mathbf{L}\mid T,G)
    =
    \prod_{j=1}^{6}
    p\!\left(
        \ell_j
        \mid
        \ell_{<j},T,G
    \right),
    \qquad
    (\ell_1,\ldots,\ell_6)
    =
    \left(
        V,\frac{b}{a},\frac{c}{a},\alpha,\beta,\gamma
    \right).
\end{equation}
With the symmetry skeleton and lattice established, the remaining problem is to localize each symmetry-inequivalent site within its admissible Wyckoff manifold.

\paragraph{Coarse-to-Fine Coordinate Generation.}
Directly predicting precise fractional coordinates is challenging because Wyckoff positions lie on manifolds of different dimensionalities and small coordinate errors can substantially alter interatomic geometry. We therefore replace one-shot continuous prediction with \textit{progressive localization}, recursively refining the feasible manifold through increasingly precise discrete spatial decisions.

For site \(i\) with Wyckoff position \(w_i\), we parameterize its fractional coordinate as $\mathbf{f}_i=\phi_{w_i}(\mathbf{u}_i), \mathbf{u}_i\in[0,1)^{d_i}$, where \(d_i=d(w_i)\) is the number of positional degrees of freedom and \(\phi_{w_i}\) maps the free parameters to the corresponding symmetry-constrained fractional coordinate. At each refinement level, we bisect every dimension of the current region in the free-parameter space of \(\mathbf{u}_i\), producing \(2^{d_i}\) candidate subregions. Thus, a general position with \(d_i=3\) yields eight candidates, a plane-constrained position with \(d_i=2\) yields four, a line-constrained position with \(d_i=1\) yields two, and a fixed position with \(d_i=0\) requires no refinement. For example, for \(w_i=(x,\frac{1}{2},\frac{1}{2})\), we have \(\mathbf{u}_i=x\) and \(d_i=1\). The first level partitions \(x\in[0,1)\) into \([0,\frac{1}{2})\) and \([\frac{1}{2},1)\); selecting the former leads the next level to refine it into \([0,\frac{1}{4})\) and \([\frac{1}{4},\frac{1}{2})\).

Extending this refinement to all symmetry-inequivalent sites, let \(\mathbf{r}_i^{(\ell)}\) denote the subregion-selection decision for site \(i\) at level \(\ell\), relative to its region at the previous level, and let
$\mathbf{R}^{(\ell)}=\{\mathbf{r}_i^{(\ell)}\}_{i=1}^{M}$. The complete refinement path is then \(\mathbf{R}^{(1:L)}=\{\mathbf{R}^{(1)},\ldots,\mathbf{R}^{(L)}\}\), whose conditional distribution is factorized level-wise as
\begin{equation}
p\left(
\mathbf{R}^{(1:L)}
\mid \mathbf{L},T,G
\right)
=
\prod_{\ell=1}^{L}
p\left(
\mathbf{R}^{(\ell)}
\mid
\mathbf{R}^{(<\ell)},
\mathbf{L},T,G
\right).
\end{equation}
Within each level, the refinement decisions for all symmetry-inequivalent sites are predicted jointly, while successive levels recursively narrow their feasible regions. After \(L\) refinements, the complete decision paths determine the free Wyckoff parameters, from which the fractional coordinates are decoded as $\mathbf{F}=\mathcal{D}\!\left(\mathbf{R}^{(1:L)},T\right)$. Thus, continuous coordinate generation is represented as a structured, symmetry-aware sequence of discrete spatial decisions.

\begin{figure*}[!t]
    \centering
    \includegraphics[width=0.99\textwidth]{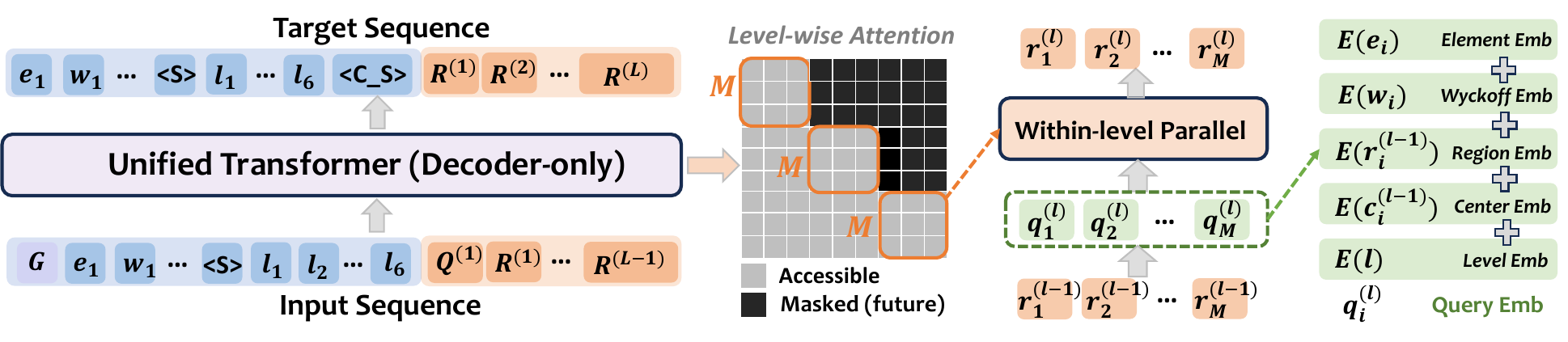}
    \caption{
    \textbf{Unified Transformer architecture of \method.} A single decoder-only Transformer jointly models the Wyckoff template, lattice, and hierarchical coordinates. Template and lattice tokens follow standard next-token prediction, while coordinates are refined level by level with causal attention across levels and parallel site interaction within each level. Each site query combines its element, Wyckoff constraint, previous region, region center, and refinement level.}
    \label{fig:CrystalAR-Figure02} 
\end{figure*}

\subsection{Unified Transformer Architecture}

The factorization above induces two dependency patterns: template and lattice variables are generated sequentially, whereas coordinate refinement jointly localizes all symmetry-inequivalent sites within each level. We therefore use a \textit{single decoder-only Transformer} that preserves the full symmetry context while transitioning from sequential crystallographic decisions to level-wise geometric refinement through block-causal attention, as illustrated in Figure~\ref{fig:CrystalAR-Figure02}.

\paragraph{From Symmetry Tokens to Geometric Queries.}
Template and lattice variables follow next-token autoregression, $G\rightarrow T\rightarrow \mathbf{L}\rightarrow
\langle\mathrm{C\_S}\rangle$. Rather than introducing a separate generation stage, the \(\langle\mathrm{C\_S}\rangle\) token initiates a set of site-specific query tokens that connect the generated symmetry context to coordinate refinement within the same decoder. For each symmetry-inequivalent site \(i\), we define a query $\mathbf{q}_i^{(\ell)}$ at refinement level \(\ell\), and collect them as $
\mathbf{Q}^{(\ell)}=\{\mathbf{q}_i^{(\ell)}\}_{i=1}^{M}$. During training, these queries are instantiated from the known site identities in the ground-truth template; during inference, they are instantiated from the previously generated template \(T=\{(e_i,w_i)\}_{i=1}^{M}\). Each query therefore carries the site's crystallographic identity together with its current localization state:
\begin{equation}
\label{eq:site_query}
    \mathbf{q}_{i}^{(\ell)}
    =
    \mathbf{E}_{\mathrm{elem}}(e_i)
    +
    \mathbf{E}_{\mathrm{wyc}}(w_i)
    +
    \mathbf{E}_{\mathrm{reg}}\!\left(R_i^{(\ell-1)}\right)
    +
    \mathbf{E}_{\mathrm{level}}(\ell)
    +
    \mathrm{PE}\!\left(\mathbf{c}_i^{(\ell-1)}\right).
\end{equation}
Here, $\mathbf{E}_{\mathrm{elem}}$ encodes the chemical identity,
$\mathbf{E}_{\mathrm{wyc}}$ specifies the Wyckoff constraint,
$\mathbf{E}_{\mathrm{reg}}$ records the region selected at the previous refinement level,
and $\mathbf{E}_{\mathrm{level}}$ distinguishes different spatial resolutions.
$\mathbf{c}_i^{(\ell-1)}$ is the decoded center of the current region, whose positional encoding provides its coarse geometric location.
For $\ell=1$, $R_i^{(0)}$ denotes the initial feasible region defined by $w_i$.


\paragraph{Level-Wise Block-Causal Attention.} The transition from discrete generation to geometric refinement requires a different attention pattern. Template and lattice tokens use standard causal attention, whereas coordinate tokens are causal only across refinement levels: sites within the same level can interact and attend to the full template, lattice, and previous refinement history, while future levels remain masked. Formally, a query at level $\ell$ may attend to ${G,T,\mathbf{L}}\cup\mathbf{R}^{(<\ell)}$ and all site queries $\mathbf{Q}^{(l)}$ at the current level. This block-causal design lets each site refine its location conditioned on the current spatial hypotheses of all others, enabling joint modeling of pairwise and higher-order geometry. Compared with site-by-site generation in CrystalFormer~\citep{CrystalFormer}, it more directly captures inter-site geometry while preserving autoregressive refinement across levels.

\paragraph{Joint Training Objective.}
All components are optimized jointly through the same Transformer. Following the autoregressive factorization above, the total training objective is
\begin{equation}
\mathcal{L}=
\lambda_{e}\mathcal{L}_{e}
+
\lambda_{w}\mathcal{L}_{w}
+
\lambda_{s}\mathcal{L}_{s}
+
\lambda_{\mathbf{L}}\mathcal{L}_{\mathbf{L}}
+
\lambda_{\mathbf{R}}\mathcal{L}_{\mathbf{R}}
\end{equation}
where $\mathcal{L}_{e}$, $\mathcal{L}_{w}$, and $\mathcal{L}_{s}$ supervise element, Wyckoff-position, and stopping-token prediction, respectively; $\mathcal{L}_{\mathbf{L}}$ supervises autoregressive lattice generation; and $\mathcal{L}_{\mathbf{R}}$ supervises hierarchical coordinate refinement across all symmetry-inequivalent sites and resolution levels. The coefficients $\lambda$ balance the relative contributions of the corresponding training objectives. Further details of the joint optimization procedure for this unified Transformer are provided in Appendix~\ref{appendix:train-inference}.

\subsection{Searching over Symmetric Crystal Space}

A key advantage of our autoregressive formulation is that crystallographic constraints can be enforced during generation. Each step defines a partial crystal state with an associated likelihood, allowing plausible hypotheses to be retained while infeasible ones are pruned. This turns crystal generation into a controllable search over progressively constrained symmetry and geometry spaces.

\paragraph{Constrained Template Search.} For CSP, let
$T_{<i}=\{(e_j,w_j)\}_{j<i}$
denote a partial template and $\mathbf{n}=\{n_e\}$ the target composition.
After generating $T_{<i}$, the remaining budget of element $e$ is
$r_e^{(i)}=n_e-\sum_{j<i}\mathbb{I}[e_j=e]\,m(w_j)$,
where $m(w_j)$ is the multiplicity of Wyckoff position $w_j$.
We restrict the next assignment $(e_i,w_i)$ to the feasible set
$\mathcal{A}(T_{<i},G,\mathbf{r}^{(i)})$, containing only choices that respect the space group and can still complete the remaining composition.
The autoregressive factorization therefore enables constrained search over
$T^{*}=\arg\max_{T\in\mathcal{T}(G,\mathbf{n})}\log p(T\mid G)$, while infeasible partial templates are eliminated before completion.
This makes template-free CSP a direct combinatorial search over valid symmetry templates rather than template retrieval followed by geometry generation, while retaining probabilistic ranking over multiple feasible crystallographic hypotheses throughout the decoding process.

\paragraph{Hierarchical Geometric Search.}
The coarse-to-fine representation induces an analogous search over geometry. A hypothesis at level $\ell$ is represented by its nested refinement path
$\mathbf{R}^{(1:\ell)}$, where
$\mathbf{R}^{(\ell)}=\{R_i^{(\ell)}\}_{i=1}^{M}$ and
$R_i^{(\ell)}\subset R_i^{(\ell-1)}$.
Each path is scored by both its autoregressive likelihood and the geometric consistency of the jointly localized sites:
\begin{equation}
S\!\left(\mathbf{R}^{(1:\ell)}\right)
=
\sum_{t=1}^{\ell}
\log p\!\left(
\mathbf{R}^{(t)}
\mid
\mathbf{R}^{(<t)},\mathbf{L},T,G
\right)
+
\lambda\,\mathcal{G}\!\left(\mathbf{R}^{(\ell)},\mathbf{L}\right),
\qquad
\mathbf{R}^{(1)}
\supset\cdots\supset
\mathbf{R}^{(\ell)} .
\end{equation}

To compute $\mathcal{G}$, we decode the center of each current region into a fractional coordinate and expand all symmetry-equivalent atoms according to the corresponding Wyckoff positions. Let $\bar{\mathbf{f}}_a^{(\ell)}$ denote the resulting fractional coordinate of atom $a$. For every distinct atom pair $(a,b)$, we compute the minimum-image Cartesian distance $d_{ab}^{(\ell)}
=
\min_{\mathbf{k}\in\mathbb{Z}^{3}}
\left\|
\mathbf{L}
\left(
\bar{\mathbf{f}}_a^{(\ell)}
-
\bar{\mathbf{f}}_b^{(\ell)}
+
\mathbf{k}
\right)
\right\|_2$ We aggregate these pairwise distances using a short-range violation penalty, $\mathcal{G}\!\left(\mathbf{R}^{(\ell)},\mathbf{L}\right)
=
-\frac{1}{|\mathcal{P}|}
\sum_{(a,b)\in\mathcal{P}}
\left[
\max\!\left(0,d_{\mathrm{min}}-d_{ab}^{(\ell)}\right)
\right]^2$,
where $\mathcal{P}$ contains all distinct symmetry-expanded atom pairs and
$d_{\mathrm{min}}$ is the minimum allowed interatomic distance.
Hypotheses with
$\min_{(a,b)\in\mathcal{P}} d_{ab}^{(\ell)}<d_{\mathrm{prune}}$
are removed immediately, while the remaining hypotheses are ranked by
$S(\mathbf{R}^{(1:\ell)})$.


\section{Experimental Results}

\begin{table*}[t]
\centering
\caption{
\textbf{De novo crystal generation on LeMat-GenBench} under \emph{with} and \emph{without pre-relaxation} settings.
We evaluate \emph{generation quality} by validity, uniqueness, and novelty;
\emph{energetic quality} by formation energy $E_f$, hull energy $E_{\mathrm{hull}}$, and relaxation RMSD;
and \emph{material discovery quality} by stability, S.U.N., metastability, and M.S.U.N.
Symmetry-native models are highlighted in red.
CrystalFormer$^{\dagger}$ is evaluated using the 1,000 publicly available generated samples, whereas the other methods use the standard evaluation set.
PLaID++ differs from the remaining baselines in that it further applies post-training to improve generation quality. \textbf{Bold} denotes the best result among symmetry-native models, while \underline{underline} denotes the best result among models without explicit template generation within each setting.
}
\label{tab:lemat_genbench}
\setlength{\tabcolsep}{3.5pt}
\renewcommand{\arraystretch}{1.08}
\begin{adjustbox}{max width=\textwidth}
\begin{tabular}{llccccccccccc}
\toprule
\multirow{2}{*}{Model}
& \multirow{2}{*}{Method}
& \multirow{2}{*}{\# Params}
& \multirow{2}{*}{Valid (\%) $\uparrow$}
& \multirow{2}{*}{Unique (\%) $\uparrow$}
& \multirow{2}{*}{Novel (\%) $\uparrow$}
& \multicolumn{3}{c}{Energy-based}
& \multicolumn{2}{c}{Stability}
& \multicolumn{2}{c}{Metastability} \\
\cmidrule(lr){7-9}
\cmidrule(lr){10-11}
\cmidrule(lr){12-13}
&
&
&
&
&
&
$E_f$ (eV) $\downarrow$
&
$E_{\mathrm{hull}}$ (eV) $\downarrow$
&
RMSD (\AA) $\downarrow$
&
Stable (\%) $\uparrow$
&
SUN (\%) $\uparrow$
&
Meta (\%) $\uparrow$
&
MSUN (\%) $\uparrow$ \\
\midrule

\multicolumn{13}{l}{\textit{With pre-relaxation}} \\

MatterGen
& Diffusion
& 47M
& 95.7
& \underline{95.1}
& \underline{70.5}
& \underline{$-0.70 \pm 0.79$}
& $0.18 \pm 0.18$
& $0.39 \pm 0.50$
& 2.0
& 0.2 (6)
& 33.4
& \underline{15.0} \\




PLaID++
& LLM+RL
& 7B
& \underline{96.0}
& 77.8
& 24.2
& $-0.50 \pm 0.44$
& \underline{$0.09 \pm 0.16$}
& \underline{$0.13 \pm 0.29$}
& \underline{12.4}
& \underline{1.0 (26)}
& \underline{60.7}
& 7.6 \\
\hdashline
\rowcolor{symred}
WyFormer
& AR+Diff
& 8B
& 93.4
& 93.0
& 66.4
& $-0.43 \pm 0.95$
& $0.50 \pm 0.51$
& $0.81 \pm 0.98$
& 0.5
& 0.1 (3)
& 15.7
& 1.9 \\
\rowcolor{symred}
WyFormer-DFT
& AR+Diff
& 8B
& \textbf{95.2}
& \textbf{95.0}
& \textbf{66.4}
& $-0.67 \pm 0.91$
& $0.27 \pm 0.36$
& $0.42 \pm 0.60$
& 3.7
& 0.4 (9)
& 24.8
& \textbf{7.8} \\
\rowcolor{symred}
\rowcolor{symred}
\method (Ours)
& AR
& 33M
& 91.4
& 89.0
& 39.3
& \textbf{-0.78 ± 0.87}
& \textbf{0.15 ± 0.26}
& \textbf{0.23 ± 0.46}
& \textbf{13.7}
& \textbf{0.9 (22)}
& \textbf{42.6}
& 7.3\\

\midrule
\multicolumn{13}{l}{\textit{Without pre-relaxation}} \\

Crystal-GFN
& GFlowNet
& 1M
& 51.7
& 51.7
& 51.7
& \underline{$-1.30 \pm 2.63$}
& $2.09 \pm 2.38$
& $1.87 \pm 0.97$
& 0.0
& 0.0 (0)
& 0.0
& 0.0 \\

ADiT
& Diffusion
& 180M
& 90.6
& 87.8
& 26.0
& $-0.73 \pm 0.93$
& $0.33 \pm 0.45$
& \underline{$0.38 \pm 0.40$}
& 0.4
& 0.0 (0)
& \underline{36.5}
& 1.0 \\

DiffCSP
& Diffusion
& 12M
& \underline{95.7}
& 94.8
& \underline{66.2}
& $-0.64 \pm 0.96$
& \underline{$0.28 \pm 0.56$}
& $0.59 \pm 0.63$
& \underline{2.3}
& 0.1 (3)
& 29.8
& \underline{8.5} \\

DiffCSP++
& Diffusion
& 12M
& 95.3
& \underline{95.1}
& 62.0
& $-0.52 \pm 0.87$
& $0.41 \pm 0.44$
& $0.69 \pm 0.82$
& 1.0
& \underline{0.2 (4)}
& 26.4
& 5.0\\




LLaMat2-CIF
& LLM
& 7B
& 84.4
& 81.4
& 30.0
& $-0.47 \pm 1.01$
& $0.44 \pm 0.71$
& $0.54 \pm 0.71$
& 0.7
& 0.1 (3)
& 34.7
& 2.1 \\

LLaMat3-CIF
& LLM
& 8B
& 15.4
& 15.2
& 10.5
& $-0.74 \pm 2.69$
& $1.71 \pm 2.48$
& $1.01 \pm 1.10$
& 0.1
& 0.0 (0)
& 2.1
& 0.2 \\
\hdashline
\rowcolor{symred}
CrystalFormer$^{\dagger}$
& AR
& 5M
& 69.9
& 69.4
& 31.8
& $-0.17 \pm 1.48$
& $\mathbf{0.70 \pm 1.33}$
& $\mathbf{0.66 \pm 1.05}$
& 1.4
& 0.0 (0)
& 28.8
& \textbf{3.1} \\


\rowcolor{symred}
SymmCD
& Diffusion
& 12M
& 73.4
& 73.0
& \textbf{47.0}
& $-0.02 \pm 1.22$
& $0.88 \pm 1.07$
& $0.87 \pm 1.09$
& 1.4
& 0.1 (2)
& 18.6
& 2.4 \\
\rowcolor{symred}
CrystalAR (Ours)
& AR
& 33M
& \textbf{91.4}
& \textbf{89.1}
& 42.8
& $\mathbf{-0.23 \pm 1.27}$
& $0.71 \pm 1.08$
& $0.77 \pm 1.25$
& \textbf{6.0}
& \textbf{0.3 (7)}
& \textbf{39.2}
& 2.7 \\

\bottomrule
\end{tabular}
\end{adjustbox}
\end{table*}

\begin{table*}[t]
  \centering
  \caption{
    \textbf{Symmetry evaluation on MP-20} using 2,500 generated structures per method.
    We group methods into models \emph{with} and \emph{without explicit Wyckoff-template generation}, with symmetry-native models highlighted in red.
    We evaluate \emph{template exploration} by the number of novel unique templates, \emph{distribution fidelity} by P1 ratio, SG $\chi^2$, $W_{N_{\mathrm{el}}}$, and $\mathrm{JSD}_{\mathrm{Wyckoff}}$, and \emph{material discovery quality} by S.U.N. and S.S.U.N.
    Symmetry statistics are computed over novel templates.
    \textbf{Bold} and \underline{underline} denote the best and second-best results, respectively. CrystalAR (SV) denotes the flexible-size variant that explicitly predicts both the number of atoms and the number of symmetry-inequivalent sites.
  }
  \resizebox{\textwidth}{!}{
  \begin{tabular}{llrrrrrrr}
    \toprule
    \multirow{2}{*}{Model}
    & \multirow{2}{*}{Method}
    & Novel Unique
    & P1 (\%)
    & \multirow{2}{*}{SG $\chi^2\downarrow$}
    & \multirow{2}{*}{$W_{N_{\mathrm{el}}}\downarrow$}
    & \multirow{2}{*}{$\mathrm{JSD}_{\mathrm{Wyckoff}}\downarrow$}
    & \multirow{2}{*}{S.U.N. (\%)$\uparrow$}
    & \multirow{2}{*}{S.S.U.N. (\%)$\uparrow$}
    \\
    &
    &
    Templates (\#)$\uparrow$
    & ref = 2.23
    &
    &
    &
    &
    \\
    \midrule

    \multicolumn{9}{l}{\textit{Models without Explicit Template Generation}} \\

    DiffCSP
    & Diffusion
    & 48 & 70.66 & 23.8303 & 0.3762 & 0.5185
    & 0.12 & 0.05 \\


    DiffCSP++
    & Diffusion
    & 5 & \textbf{2.19} & \textbf{0.1049} & 0.4787 & 0.0343
    & \underline{0.19} & \underline{0.19} \\

    \hdashline
    \rowcolor{symred}
    \multicolumn{9}{l}{\textit{Symmetry-native Models}} \\

    \rowcolor{symred}
    WyFormer$_{\mathrm{DiffCSP++}}$
    & AR + Diffusion
    & 190 & 1.56 & 0.2238 & 0.1087 & \underline{0.0248}
    & 0.13 & 0.13 \\
    
    \rowcolor{symred}
    SGEquiDiff
    & AR + Diffusion
    & \textbf{386} & 2.02 & 0.2112
    & 0.1641 & \textbf{0.0218}
    & 0.11 & 0.11 \\

    \rowcolor{symred}
    SymmCD
    & Diffusion
    & 215 & 2.38 & \underline{0.1276} & 0.5116
    & 0.0290 & 0.09 & 0.09 \\

    \rowcolor{symred}
    CrystalFormer
    & Autoregressive
    & 89 & 1.26 & 0.3138 & 0.1697 & 0.0399
    & 0.00 & 0.00 \\

    \rowcolor{symred}
    CrystalAR (SV)
    & Autoregressive
    & \underline{268} & 1.94 & 0.2584
    & \underline{0.0973} & 0.0323
    & \underline{0.19} & \underline{0.19} \\

    \rowcolor{symred}
    CrystalAR (Ours)
    & Autoregressive
    & 161 & \underline{2.42} & 0.3469
    & \textbf{0.0611} & 0.0414
    & \textbf{0.28} & \textbf{0.28} \\

    \bottomrule
  \end{tabular}
  }
  \label{tab:mp20_raw_generation}
\end{table*}

\subsection{Experimental Setup} 

\paragraph{Tasks.} We consider two complementary tasks: \textbf{de novo crystal generation} and \textbf{crystal structure prediction (CSP)}. For de novo generation, we follow standard benchmarks to evaluate general material-discovery quality, including stable, unique, and novel (S.U.N.) structures. We further introduce a symmetry-aware evaluation that measures adherence to prescribed crystallographic symmetry and exploration of unseen Wyckoff templates. Template novelty is stricter than structure-level novelty, since a structurally novel crystal may still reuse a template seen during training, and thus better reflects whether models discover genuinely new crystallographic configurations. For \textbf{CSP}, we evaluate whether the model can recover the ground-truth structure from a given composition, providing a stringent test of geometric precision.

\paragraph{Datasets \& Baselines.} We conduct experiments on two widely used crystal-generation benchmarks, \textbf{MP-20} and \textbf{MPTS-52}. MP-20 contains experimentally observed inorganic crystals with at most 20 atoms in the primitive cell and serves as our primary benchmark for both tasks, while MPTS-52 contains larger and more structurally complex crystals and is used for CSP to evaluate scalability and generalization. We compare \method with two broad classes of methods. The first consists of \emph{symmetry-native} approaches that explicitly model crystallographic templates, including WyFormer~\citep{WyFormer}, SGEquiDiff~\citep{SGEquiDiff}, CrystalFormer~\citep{CrystalFormer}, and SymmCD~\citep{SymmCD}; \method also falls into this category. The second includes more general crystal generative models without explicit template generation, including MatterGen~\citep{MatterGen}, PLaiD~\citep{plaid}, Crystal-GFN~\citep{crystal-GFN}, ADiT~\citep{ADIT}, DiffCSP~\citep{DiffCSP}, DiffCSP++~\citep{DiffCSP++}, LLaMat-CIF~\citep{llamat-CIF}, FlowMM~\citep{FlowMM}, CDVAE~\citep{CDVAE}, CrystalFlow~\citep{CrystalFlow}, and OMatG~\citep{OMG}. For each setting, we select task-appropriate baselines and restrict comparisons to methods with \textit{publicly available implementations} for reproducibility.

\begin{figure*}[!t]
    \centering
    \includegraphics[width=0.99\textwidth]{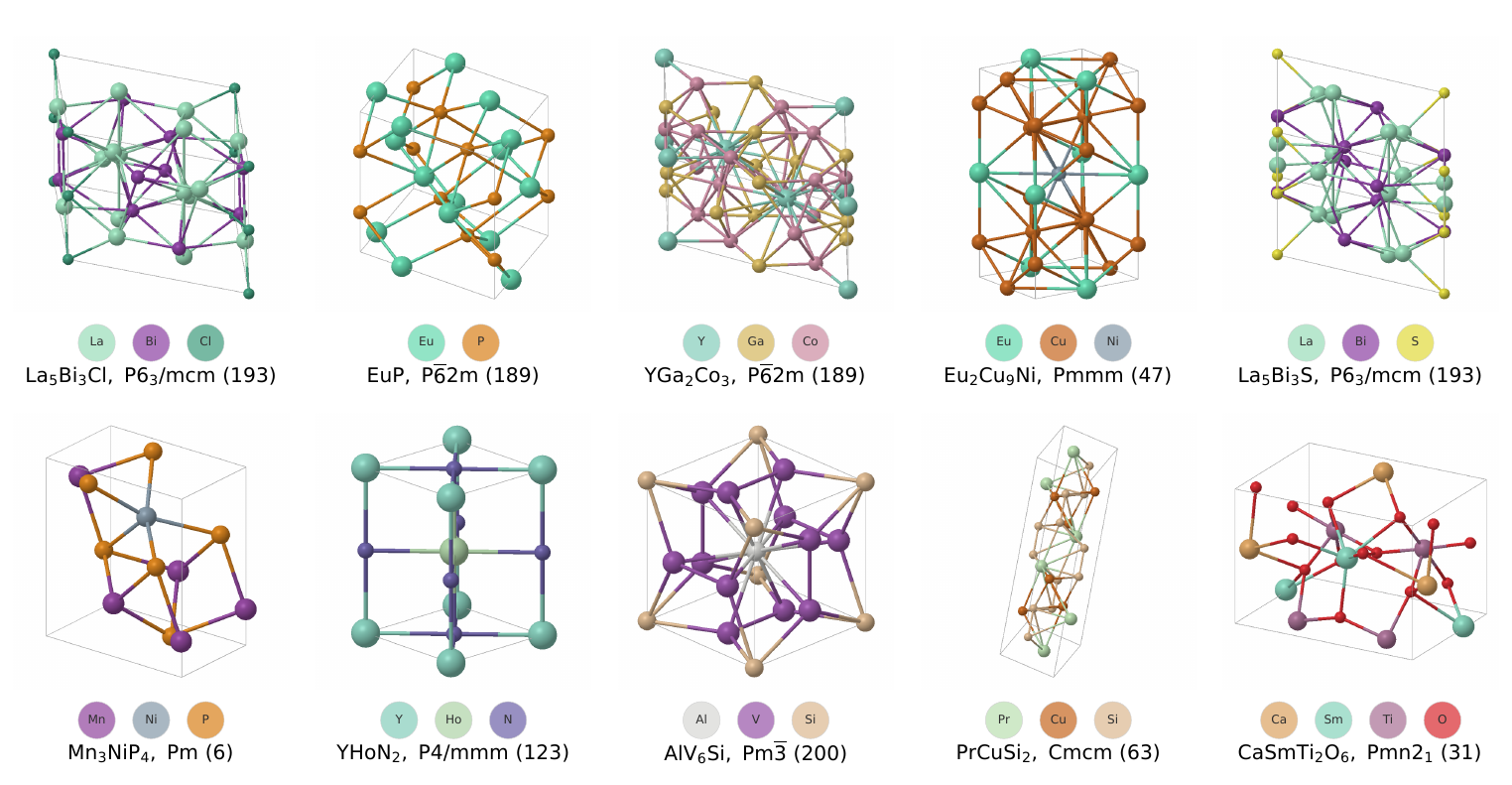}
    \caption{Representative S.S.U.N. crystal structures generated by \method, illustrating diverse chemical compositions, space groups, and crystallographic configurations.} 
    \label{fig:CrystalAR-Figure03} 
\end{figure*}

\subsection{De novo Generation}

Table~\ref{tab:lemat_genbench} evaluates whether explicit symmetry modeling can maintain strong material-discovery performance on the standard de novo generation benchmark. Existing symmetry-native methods show a noticeable gap from models without explicit template generation, particularly on stability-oriented metrics, suggesting that enforcing crystallographic symmetry alone is insufficient for high-quality generation. \method substantially narrows this gap. Without pre-relaxation, it achieves a stability rate of 6.0\% and an S.U.N. rate of 0.3\%, compared with 1.4\%/0.0\% for CrystalFormer and 1.4\%/0.1\% for SymmCD. These results are already competitive with strong models without explicit template generation, whose best reported stability and S.U.N. rates are 2.3\% and 0.2\%, respectively. After pre-relaxation, \method further improves to 13.7\% stability and 0.9\% S.U.N., substantially outperforming WyFormer-DFT (3.7\% and 0.4\%) and approaching the post-trained PLaID++ (12.4\% and 1.0\%). \method also achieves the strongest energetic quality among symmetry-native models, with $E_f=-0.78$ eV and $E_{\mathrm{hull}}=0.15$ eV after pre-relaxation. Overall, these results show that symmetry-native generation need not trade material quality for explicit crystallographic modeling: \method brings material-discovery performance into the regime of broader crystal generators while preserving explicit symmetry structure, establishing symmetry-native generation as a competitive paradigm for practical materials discovery.

\subsection{Symmetry-Aware Evaluation for Crystal Generation}

Table~\ref{tab:mp20_raw_generation} provides a stricter symmetry-aware view of de novo generation by evaluating not only whether generated structures are novel, but whether they arise from genuinely novel crystallographic templates. This distinction is important because structural novelty can still be achieved within symmetry patterns already seen during training, whereas template novelty requires exploration of previously unseen Wyckoff configurations. Models without explicit template generation show clear limitations under this criterion: DiffCSP discovers only 48 novel unique templates and exhibits large deviations from the reference space-group and Wyckoff distributions, while DiffCSP++ closely matches the symmetry distribution but discovers only five novel templates. In contrast, symmetry-native models substantially expand the accessible template space. \method (SV) discovers 268 novel unique templates, exceeding SymmCD (215), while maintaining strong symmetry fidelity, with a P1 ratio of 1.94 (reference: 2.23), $W_{N_{\mathrm{el}}}=0.0973$, and $\mathrm{JSD}_{\mathrm{Wyckoff}}=0.0323$. More importantly, this broader template exploration remains effective for material discovery: \method (SV) reaches 0.19\% S.S.U.N., more than doubling SymmCD (0.09\%). In comparison, the default \method samples atom and site counts from the empirical joint distribution, yielding fewer novel templates but higher S.U.N. and S.S.U.N. rates of 0.28\%. This comparison highlights a size-modeling trade-off: sampling atom and site counts from the empirical joint distribution provides a stronger stability prior, whereas explicitly generating both quantities broadens the explored size space and improves template novelty.

\subsection{Crystal Structure Prediction (CSP)} 

\begin{wraptable}{r}{0.60\textwidth}
\vspace{-1.8\intextsep}
\centering
\caption{
\textbf{Crystal structure prediction on MP-20 and MPTS-52} under single-hypothesis (1) and multi-hypothesis (20) generation.
We evaluate \emph{structure recovery} by match rate (MR, $\uparrow$) and \emph{geometric accuracy} by RMSE ($\downarrow$).
\method performs template-free CSP without supplying a retrieved Wyckoff template.
\textbf{Bold} denotes the best result in each column, while \underline{underline} denotes the best single-hypothesis result.
}
\label{tab:csp_results}
\resizebox{\linewidth}{!}{
\begin{tabular}{lccccc}
\toprule
& \multicolumn{2}{c}{MP-20}
& \multicolumn{2}{c}{MPTS-52} \\
\cmidrule(lr){2-3}
\cmidrule(lr){4-5}
Model
& MR (\%) $\uparrow$
& RMSE $\downarrow$
& MR (\%) $\uparrow$
& RMSE $\downarrow$ \\
\midrule
CDVAE (1)
& 33.90 & 0.1045
& 5.34 & 0.2106 \\

DiffCSP (1)
& 51.49 & 0.0631
& 12.19 & 0.1786 \\

FlowMM (1)
& 61.39 & \textbf{0.0566}
& 17.54 & 0.1726 \\

CrystalFlow (1)
& 62.02 & 0.0710
& 21.00 & \underline{0.1613} \\

OMatG (1)
& \textbf{63.75} & 0.0720
& \underline{25.15} & 0.1931 \\
\hdashline
\rowcolor{symred}
\method{} (1)
& \underline{62.92} & \underline{0.0585}
& \textbf{25.25} & \textbf{0.1305} \\
\rowcolor{symred}
CrystalFormer (1)
& 29.62 & 0.0619
& - & - \\
\rowcolor{symred}
WyFormer (1)
& 10.25 & 0.0522 (NA)
& - & - \\



\midrule




DiffCSP (20)
& \underline{77.93} & \textbf{0.0492}
& 34.02 & 0.1749 \\

CrystalFlow (20)
& \textbf{78.34} & \underline{0.0577}
& \textbf{37.81} & \underline{0.1584} \\
\hdashline
\rowcolor{symred}
\method{} (20)
& 74.55 & 0.0655
& \underline{35.71} & \textbf{0.1454} \\
\rowcolor{symred}
CrystalFormer (20)
& 64.56 & 0.0835
& - & - \\
\rowcolor{symred}
WyFormer (20)
& 40.90 & 0.0472 (NA)
& - & - \\

\bottomrule
\end{tabular}
}
\end{wraptable}
We next evaluate \method on \emph{template-free} CSP, where only the chemical composition and space group are provided and the Wyckoff template must be inferred by the model. This setting directly tests whether symmetry-native models can recover both the latent Wyckoff configuration and the corresponding geometry without relying on template retrieval. SGEquiDiff and SymmCD are excluded because their formulations do not support template-free CSP inference; ``--'' denotes settings for which results are not reported due to prohibitively long inference time. As shown in Table~\ref{tab:csp_results}, \method achieves a single-hypothesis match rate of 62.92\% with an RMSE of 0.0585 on MP-20, closely matching the strongest diffusion- and flow-based CSP models while substantially outperforming previous symmetry-native approaches. The advantage is more pronounced on the larger MPTS-52 benchmark, where \method achieves the best single-hypothesis results in both structure recovery (25.25\% MR) and geometric accuracy (0.1305 RMSE). Multi-hypothesis decoding further improves the match rate to 74.55\% on MP-20 and 35.71\% on MPTS-52 by exploring alternative symmetry-compatible structures. These results show that \method can retain explicit Wyckoff-template generation while achieving competitive geometric precision in template-free CSP, including on more complex crystal structures. “NA” denotes results excluded from the ranking due to the low MR.

\section{Conclusion}

We present \method, a unified autoregressive framework for symmetry-native crystal generation. \method factorizes generation into Wyckoff templates, lattice parameters, and symmetry-constrained coordinates, following the hierarchical dependencies imposed by crystallographic symmetry. Its level-wise coarse-to-fine representation further casts continuous coordinate generation as structured search over progressively refined Wyckoff-constrained regions. Across symmetry-constrained generation, de novo generation, and crystal structure prediction, \method balances template novelty and structural stability while improving geometric accuracy over prior symmetry-native methods. It achieves competitive template-free CSP without retrieved Wyckoff templates and brings symmetry-native de novo generation to a performance level comparable to general crystal generators. These results highlight autoregressive crystallographic search as a unified paradigm for exploring novel symmetry configurations and high-quality crystal structures.

\section*{AI Use Statement}

In this work, we used generative AI tools to assist with writing and language polishing. Generative AI was not used for reference retrieval, and automated research tools were not used in this work. We additionally used generative AI to assist with method implementation. All AI-assisted outputs, including implementation support and writing suggestions, were carefully reviewed and verified by the authors. The authors take full responsibility for the final content of this work, including all text, claims, implementations, and other artifacts produced with the assistance of generative AI.

\section*{Ethics Statement}

This work develops generative models for inorganic crystal structures and does not involve human subjects, personal data, or other sensitive information. The proposed approach is designed to advance computational materials discovery by enabling more effective exploration of symmetry-constrained crystal structures. As with other generative materials models, predicted structures should undergo appropriate computational and experimental validation before practical deployment. We do not identify any direct ethical concerns associated with the datasets, methods, or experiments used in this study.

\section*{Reproducibility Statement}

We benchmark \method on widely used crystal-generation datasets, including MP-20, MPTS-52, and LeMat-GenBench, following established evaluation protocols where applicable. The manuscript provides the key methodological details needed for reproduction, including the autoregressive factorization, symmetry-constrained coordinate modeling, decoding strategy, and evaluation metrics. We plan to release the implementation, model configurations, trained checkpoints, generation scripts, and evaluation code, together with the generated structures reported in our experiments. Preprocessing procedures and benchmark-specific decoding settings will also be documented.

\section*{Acknowledgment}

This research/project is supported by the National Research Foundation, Singapore under its AI Singapore Programme (AISG Award No: AISG3-RP-2022-028).

\newpage
\bibliography{iclr2027_conference}
\bibliographystyle{iclr2027_conference}

\newpage

\appendix
\section{Appendix}

\subsection{Lattice Volume--Shape Decomposition.} 
\label{appendix:lattice}

We transform the lattice matrix $\mathbf{L}\in\mathbb{R}^{3\times 3}$ into a six-dimensional representation that explicitly separates the unit-cell volume from its shape. Let $\mathbf{a},\mathbf{b},\mathbf{c}$ denote the three lattice vectors, with lengths $a=\|\mathbf{a}\|$, $b=\|\mathbf{b}\|$, and $c=\|\mathbf{c}\|$. We decompose the lattice as
\[
\mathbf{L}
\;\longleftrightarrow\;
\boldsymbol{\ell}
=
\left(
V,\,
r_b,\,
r_c,\,
\alpha,\,
\beta,\,
\gamma
\right),
\]
where
\[
V = |\det(\mathbf{L})|,
\qquad
r_b = \frac{b}{a},
\qquad
r_c = \frac{c}{a},
\]
and $\alpha=\angle(\mathbf{b},\mathbf{c})$,
$\beta=\angle(\mathbf{a},\mathbf{c})$, and
$\gamma=\angle(\mathbf{a},\mathbf{b})$ are the three lattice angles.
Here, $V$ captures the overall scale of the unit cell, while
$(r_b,r_c,\alpha,\beta,\gamma)$ characterize its shape independently of
absolute size. The transformation is invertible: defining
\[
q =
\sqrt{
1+2\cos\alpha\cos\beta\cos\gamma
-\cos^2\alpha-\cos^2\beta-\cos^2\gamma
},
\]
the lattice lengths are recovered as
\[
a=\left(\frac{V}{r_b r_c q}\right)^{1/3},
\qquad
b=r_ba,
\qquad
c=r_ca,
\]
after which the lattice matrix can be reconstructed under the standard
crystallographic convention.

\subsection{More Algorithm Details}

\paragraph{Stopping Criterion.} The stopping rule depends on how the template size is specified. For the SV variant, the model directly predicts the atom and symmetry-inequivalent site counts, and template decoding terminates once the predicted number of sites has been generated. For the non-SV variant of \method, these size variables are sampled from the empirical joint distribution of space group, symmetry-inequivalent site count, atom count and element count $(G,M,N,K)$, estimated from the training data. Given the sampled size variables, decoding terminates once the corresponding site count is reached. In the \emph{free} setting, no template size is provided in advance. During training, we append a \textsc{STOP} token $\langle S\rangle$ to the end of each template sequence, so that the model learns to infer the template length autoregressively. At inference time, following a strategy similar to SGEquiDiff and WyFormer, generation terminates once the model predicts this binary \textsc{STOP} action. For CSP, the prescribed composition provides a deterministic stopping criterion: decoding ends once the multiplicity-weighted site assignments exactly exhaust the target elemental counts.

\paragraph{Coordinate Decoding.}
Each coordinate token \(r_i^{(\ell)}\) specifies a child region relative to the region selected at the previous refinement level. Starting from the full fractional-coordinate cell \(\mathcal{C}_i^{(0)}=[0,1)^3\), the complete refinement path recursively determines $\mathcal{C}_i^{(\ell)}
=
\operatorname{Child}\!\left(
\mathcal{C}_i^{(\ell-1)},r_i^{(\ell)}
\right),
\qquad
\ell=1,\ldots,L$ Only child cells intersecting the Wyckoff manifold of \(w_i\) are admissible. After \(L\) refinements, the complete path \(r_i^{(1:L)}\) therefore specifies a finest feasible region on this manifold. We recover the corresponding free Wyckoff parameters from this region and map them through the Wyckoff parameterization, $\hat{\mathbf f}_i
=
\phi_{w_i}\!\left(
\hat{\mathbf u}_i
\right),
$ yielding a fractional coordinate that satisfies the prescribed site symmetry by construction. Applying this procedure to all inequivalent sites gives \(\mathbf F=\mathcal{D}(\mathbf R^{(1:L)},T)\); the full crystal is then obtained by applying the symmetry operations of \(G\).

\paragraph{Token Embedding.} All input variables are embedded into a common \(D\)-dimensional hidden space before entering the Transformer. Discrete crystallographic variables use separate learned lookup tables: the space group is represented by \(\mathbf{E}_{G}(G)\), the element identity by \(\mathbf{E}_{\mathrm{elem}}(e_i)\), and the Wyckoff information by \(\mathbf{E}_{\mathrm{wyc}}(w_i)\), which encodes the corresponding site-symmetry type and coordinate-form index. Special control tokens, such as \(\langle\mathrm{C\_S}\rangle\), are assigned independent learned embeddings. Continuous lattice variables are first normalized according to the crystal family and then projected into the hidden space together with embeddings indicating their positions in the lattice-generation sequence. For coordinate refinement, we use the site-specific query embedding defined in Eq.~\ref{eq:site_query}. Here, \(\mathbf{E}_{\mathrm{elem}}(e_i)\) and \(\mathbf{E}_{\mathrm{wyc}}(w_i)\) preserve the chemical and crystallographic identity of site \(i\); \(\mathbf{E}_{\mathrm{reg}}(r_i^{(\ell-1)})\) embeds the child-region decision from the previous refinement level; and \(\mathbf{E}_{\mathrm{level}}(\ell)\) identifies the current spatial resolution. The previous refinement path determines the current feasible region, whose decoded center \(\mathbf{c}_i^{(\ell-1)}\in[0,1)^3\) is further encoded by \(\mathrm{PE}(\mathbf{c}_i^{(\ell-1)})\) to provide its coarse spatial location. For \(\ell=1\), \(r_i^{(0)}\) and \(\mathbf{c}_i^{(0)}\) are initialized from the feasible region specified by \(w_i\). Together, these embeddings allow each query to carry both the fixed symmetry context and the evolving geometric state throughout hierarchical refinement.

\paragraph{Variable Prediction Heads.}
\method employs variable-specific output heads on top of the shared Transformer representation, using categorical projections for discrete crystallographic decisions and conditional density heads for continuous lattice variables. The hidden width is \(D=384\) for MP-20 and \(D=512\) for MPTS-52. For each discrete variable \(v\), an affine projection $\mathbf z_v=\mathbf W_v\mathbf h+\mathbf b_v$ produces its logits, with inadmissible actions masked to \(-\infty\) before normalization. Template generation predicts the element, site-symmetry type, and coordinate-form index using shared categorical heads with 89, 75, and 8 outputs, respectively; together with the space group, the latter two identify the corresponding Wyckoff orbit.

Lattice variables are predicted using conditional neural spline flow (NSF) heads. We first transform and standardize the lattice according to its crystal family, leaving \(1,2,3,4,\) or \(6\) active variables for cubic, tetragonal/hexagonal, orthorhombic, monoclinic, and triclinic systems, respectively. Each active scalar is generated autoregressively by a separate NSF conditioned on the Transformer state and previously generated lattice variables. The NSF parameter network has dimensions $(D+6)\rightarrow256\rightarrow256\rightarrow23$, where the 23 outputs parameterize an eight-bin rational-quadratic spline. Inverse normalization and the family-specific geometric decoder subsequently recover the physical lattice parameters while preserving the corresponding lattice symmetry.

For coordinate refinement, a single shared linear head $\mathbb{R}^{D}\rightarrow\mathbb{R}^{8}$ predicts the eight octree child cells for every inequivalent site and refinement level. The same head is shared across all sites and all \(L=9\) levels, with a Wyckoff-dependent mask removing child cells incompatible with the site's symmetry manifold. The model further includes task-specific auxiliary heads for CSP multiplicity prediction, orbit-count prediction, and composition-conditioned space-group, crystal-family, and centering-type prediction.

\subsection{Training \& Inference}
\label{appendix:train-inference}

The following pseudocode makes explicit the two training paths and the two
inference paths.  In all four algorithms,
$T=(A,W)=\{(e_i,w_i)\}_{i=1}^{M}$ is the element-decorated Wyckoff template,
$\widetilde{\mathbf y}$ is the family-normalized lattice vector, and
$\mathbf R^{(1:L)}$ is the level-wise coordinate-refinement path. Algorithms~\ref{alg:csp-training} and~\ref{alg:csp-inference} present the detailed training and inference procedures for CSP, respectively, while Algorithms~\ref{alg:denovo-training} and~\ref{alg:denovo-inference} describe the corresponding procedures for de novo crystal generation.

\begin{algorithm}[t]
\caption{CSP training of \method}
\label{alg:csp-training}
\begin{algorithmic}[1]
\State \textbf{Input:} CSP training set
\(\mathcal D_{\mathrm{CSP}}=\{(\mathbf n,G,C)\}\),
pretrained parameters $\theta_{\mathrm{base}}$
\State \textbf{Initialize:} freeze $\theta_{\mathrm{base}}$ and attach a
rank-$32$ adapter $\theta_{\mathrm{CSP}}$ and orbit-count head $q_\psi(M\mid\mathbf n,G)$
\While{not converged}
    \State Sample $(\mathbf n,G,C)$ and symmetry-decompose $C$ into
    $T^\star=(A^\star,W^\star)$, $\widetilde{\mathbf y}^{\star}$, and
    $\mathbf R^{\star(1:L)}$
    \For{$i=1,\ldots,M$}
        \State Compute remaining composition
        $r_e^{(i)}=n_e-\sum_{j<i}\mathbb I[e_j=e]m(w_j)$
        and feasible actions
        $\mathcal A_i=\mathcal A(T_{<i},G,\mathbf r^{(i)})$
    \EndFor
    \State Teacher-force $(G,\mathbf n,T^\star,\widetilde{\mathbf y}^{\star},
    \mathbf R^{\star(1:L)})$ with causal template/lattice attention and
    level-wise block-causal coordinate attention
    \State Evaluate masked template loss
    $\mathcal L_T^{\mathcal A}$ only over $\mathcal A_i$, lattice loss
    $\mathcal L_{\mathbf{L}}$, coordinate region loss
    $\mathcal L_\mathbf{R}=\sum_{\ell=1}^{L}\mathcal L_\mathbf{R}^{(\ell)}$, and
    orbit-count loss $\mathcal L_M$
    \State Update only $(\theta_{\mathrm{CSP}},\psi)$ using
    $\displaystyle\mathcal L_{\mathrm{CSP}}=
    2\lambda_T\mathcal L_T^{\mathcal A}+
    \lambda_\mathbf{L}\mathcal L_\mathbf{L}+\lambda_\mathbf{R}\mathcal L_\mathbf{R}+
    \lambda_M\mathcal L_M$
\EndWhile
\State \textbf{Output:} $\theta_{\mathrm{CSP}}$ for composition- and
space-group-conditioned inference
\end{algorithmic}
\end{algorithm}

\begin{algorithm}[t]
\caption{De novo training of \method}
\label{alg:denovo-training}
\begin{algorithmic}[1]
\State \textbf{Input:} Crystal dataset $\mathcal D$; model parameters $\theta$
\While{not converged}
    \State Sample $C=(G,\mathbf n,\mathbf L,T,\mathbf F)$ from $\mathcal D$
    and record $(N,M,K)=(\sum_e n_e,|T|,|\{e:e\in\mathbf n\}|)$
    \State Encode $\mathbf L$ as $\widetilde{\mathbf y}$ and encode the
    free Wyckoff coordinates as $\mathbf R^{\star(1:L)}$
    \State Construct the teacher-forced sequence
    $(G,T^\star,\widetilde{\mathbf y}^{\star},\mathbf R^{\star(1:L)})$;
    use the Wyckoff-manifold mask for every coordinate level
    \State Predict $e_i\rightarrow w_i$ autoregressively, the six active
    lattice variables with NSF heads, and all sites' child regions jointly
    at each level
    \State Compute $\mathcal L_T=\mathcal L_e+\mathcal L_w+\mathcal L_{\mathrm{stop}}$,
    $\mathcal L_\mathbf{L}$, and $\mathcal L_\mathbf{R}=\sum_{\ell=1}^{L}\mathcal L_\mathbf{R}^{(\ell)}$
    \State Update all parameters with
    $\displaystyle\mathcal L=\lambda_T\mathcal L_T+
    \lambda_\mathbf{L}\mathcal L_\mathbf{L}+\lambda_\mathbf{R}\mathcal L_\mathbf{R}+
    \lambda_{\mathrm{aux}}\mathcal L_{\mathrm{aux}}$
\EndWhile
\State \textbf{Output:} $\theta$ and the empirical prior
 $p_{\mathrm{train}}(G,N,M,K)$ used for MP-20 de novo sampling
\end{algorithmic}
\end{algorithm}

\begin{algorithm}[t]
\caption{De novo inference with simulation-guided beam search}
\label{alg:denovo-inference}
\begin{algorithmic}[1]
\State \textbf{Input:} $\theta$, $p_{\mathrm{train}}(G,N,M,K)$, and
number of attempts $S=2500$
\For{$s=1,\ldots,S$}
    \State Sample $(G,N,M,K)\sim p_{\mathrm{train}}$; generate $T$ and
    $\widetilde{\mathbf y}$ autoregressively with temperatures
    $(\tau_T,\tau_L)=(1,1)$
    \State Initialize the coordinate beam
    $\mathcal B_0=\{\varnothing\}$ and prefix budget $Q=2049$
    \For{$\ell=1,\ldots,L=9$}
        \State Expand every prefix in $\mathcal B_{\ell-1}$ to at most
        $E=56$ masked child assignments for all sites jointly, using
        coordinate temperature $\tau_F=0.7$
        \State Greedily complete each expanded prefix to level $L$;
        apply Wyckoff masks and symmetry-expand each completed path to
        $\widehat C(\mathbf R^{(1:L)})$
        \State For every complete rollout, compute capped clearance
        $c(\pi)=\min\{c_{\max},\operatorname{clr}(\widehat C(\pi))\}$
        from minimum-image interatomic distances, with
        $c_{\max}=1.2899$
        \State Rank complete rollouts by $(c(\pi),\log p_\theta(\pi))$;
        retain the top beam $B=4$ prefixes and charge every model prefix
        evaluation to the global budget $Q$
        \If{$\mathcal B_\ell=\varnothing$ or the budget is exhausted}
            \State \textbf{break}
        \EndIf
    \EndFor
    \State Decode the best $\mathbf F$ from the retained path and apply
    $G$'s symmetry operations to obtain the full crystal $\widehat C_s$
\EndFor
\State Optionally apply DiffCSP++ pre-relaxation and evaluate the three-MLIP
    mean energy; retain failed attempts in the denominator
\State \textbf{Output:} $\{\widehat C_s\}_{s=1}^{2500}$
\end{algorithmic}
\end{algorithm}

\begin{algorithm}[t]
\caption{Template-free CSP inference with template and coordinate search}
\label{alg:csp-inference}
\begin{algorithmic}[1]
\State \textbf{Input:} composition $\mathbf n$, space group $G$,
trained $\theta_{\mathrm{CSP}}$, and hypothesis count $K\in\{1,20\}$
\State Initialize a template beam $\mathcal B_T^0=\{\varnothing\}$
\For{$i=1,\ldots,M$}
    \State Expand each partial template with local width $4$ using only
    $(e_i,w_i)\in\mathcal A(T_{<i},G,\mathbf r^{(i)})$
    \State Keep the top $32$ partial templates by model score
\EndFor
\State Complete and rank feasible templates by
    $s_T(T)=\log p_\theta(T\mid\mathbf n,G)+\log q_\psi(M\mid\mathbf n,G)$;
cycle through the available templates only if fewer than $K$ are complete
\For{$j=1,\ldots,K$}
    \State Select ranked template $T_j$ and sample
    $\widetilde{\mathbf y}_j\sim p_\theta(\widetilde{\mathbf y}\mid T_j,\mathbf n,G)$
    with lattice seed $j+1$; initialize $\mathcal B_0=\{\varnothing\}$
    \For{$\ell=1,\ldots,L=9$}
        \State Expand each prefix to at most $32$ child paths; mask
        Wyckoff-incompatible children and any periodic contact below
        $0.5$~\AA{}; spend at most $Q=512$ global prefix evaluations
        \State Greedily roll out each inspected prefix to level $L$,
        symmetry-expand it, and compute capped clearance
        $c(\pi)=\min\{1.2899,\operatorname{clr}(\widehat C(\pi))\}$
        \State If $T_j$ has repeated free orbits, rank by
        $\log p_\theta(\pi)+8c(\pi)$; otherwise rank lexicographically by
        $(c(\pi),\log p_\theta(\pi))$; retain the top $16$ prefixes
    \EndFor
\State Decode $\mathbf F_j$ from the best complete path and construct
    $\widehat C_j=\operatorname{Expand}_G(T_j,\mathbf L_j,\mathbf F_j)$
\EndFor
\State \textbf{Output:} $K$ symmetry-consistent crystal structures
\end{algorithmic}
\end{algorithm}

\subsection{More Related Work Discussion}

\paragraph{Generative models} A variety of generative paradigms have been explored for crystal generation. Early studies employed variational autoencoders~\citep{vae,WyCryst} and generative adversarial networks~\citep{gan,crystalgan} to model the distribution of crystalline structures. More recently, diffusion models~\citep{Diffusion,DDIM,Score,ScoreSDE,DiffCSP,DiffCSP++,UniMat} and flow-matching approaches~\citep{FlowMM,Reflow,mean-flow} have emerged as dominant frameworks, achieving strong performance in both crystal generation and structure prediction. Related continuous-time formulations, such as stochastic interpolants~\citep{sto-interpolant,OMG}, have also been explored as flexible alternatives within this family of generative methods. By contrast, autoregressive modeling~\citep{seq2seq}, despite its remarkable success in language modeling~\citep{GPT-3,GPT-4}, remains comparatively underexplored for crystal generation. Several recent works have directly adapted large language models to crystal generation~\citep{NatureLM,CrystalLLM,UniGenX}; however, such sequence-based formulations often face difficulties in effectively exploring novel crystal structures and in modeling precise continuous geometry. In this work, we instead adopt a coarse-to-fine autoregressive formulation for continuous geometric generation, inspired by hierarchical autoregressive paradigms developed for image generation~\citep{VAR}, video generation~\citep{infinitystar} and 3D point-cloud modeling~\citep{pointnsp,octgpt,octree-transformer}. 

\paragraph{More Crystal Generation Works} Beyond the approaches discussed above, several other directions have also been explored for crystal generation. CrysBFN~\citep{crys-bfn} introduces Bayesian flow networks~\citep{BFN} to crystal generation, while SymmBFN~\citep{symmbfn} further incorporates crystallographic symmetry into this framework. PXRD~\citep{pxrd} studies conditional crystal generation from powder X-ray diffraction patterns. More recently, several works~\citep{park2026guiding,plaid} have begun to investigate post-training strategies, including preference-optimization methods such as DPO~\citep{dpo} and GRPO~\citep{GRPO}, to further improve the quality and discovery capability of crystal generative models. Another emerging direction focuses on the generation of disordered crystal structures~\citep{disgen,dmflow}, extending generative modeling beyond perfectly ordered crystalline systems.

\begin{table*}[ht!]
\centering
\caption{``Generative Model” denotes the underlying generative modeling backbone employed by each method. ``Unified” indicates whether the model is trained and performs inference over the complete crystal geometry in an end-to-end fashion using a single unified model. ``Symmetry” specifies whether Wyckoff symmetry is explicitly modeled at the level of individual atomic sites. ``Template-Gen” indicates whether Wyckoff positions are generated by the model itself, rather than retrieved from predefined symmetry templates or sampled from the empirical distribution of the training data. ``Template-Free CSP” indicates whether the method supports crystal structure prediction without template retrieval. ``Inter-Geom” indicates whether the model explicitly and effectively captures interatomic geometry. ``-” denotes that the item is not applicable.} 
\scalebox{0.70}{
\begin{tabular}{lcccccc} 
\toprule
\textbf{Model} & Generative Model & Unified & Symmetry & Template-Gen & Template-free CSP & Inter-Geom \\
\midrule
CDVAE~\citep{CDVAE} & VAE+Diff. & \checkmark & &  & \checkmark & \checkmark\\ 
DiffCSP~\citep{DiffCSP} & Diff. & \checkmark & &  & \checkmark & \checkmark \\
MatterGen~\citep{MatterGen} & Diff. & \checkmark & &  & \checkmark & \checkmark \\
KLDM~\citep{KLDM} & Diff. & \checkmark & &  & \checkmark & \checkmark \\ 
ADiT~\citep{ADIT} & Diff. & \checkmark & &  & & \checkmark \\
FlowMM~\citep{FlowMM} & Flow & \checkmark & &  & \checkmark & \checkmark \\
FlowLLM~\citep{FlowLLM} & LLM+Flow & \checkmark & &  & & \checkmark\\
CrystalFlow~\citep{CrystalFlow} & Flow & \checkmark & &  & \checkmark & \checkmark\\
OMatG~\citep{OMG} & Flow & \checkmark & &  & \checkmark & \checkmark \\ 
MCFlow~\citep{MCFlow} & Flow & \checkmark & &  & \checkmark & \checkmark \\ 
\hline
WyCrys~\citep{WyCryst} & VAE & & \checkmark & \checkmark & & \\
DiffCSP++~\citep{DiffCSP++} & Diff. & \checkmark & \checkmark & & \checkmark & \checkmark \\
WyckoffDiff~\citep{WyckoffDiff} & Diff. & & \checkmark & \checkmark & & \\ 
SGEquiDiff~\citep{SGEquiDiff} & AR+Diff & & \checkmark & \checkmark &  & \checkmark \\
WyFormer~\citep{WyFormer} & AR+Diff & & \checkmark & \checkmark & &  \\
SGFM~\citep{SGFM} & Flow & \checkmark & \checkmark & &  & \checkmark \\
SymmCD~\citep{SymmCD} & Diff. & \checkmark & \checkmark & \checkmark & & \checkmark \\
CrystalFormer~\citep{CrystalFormer} & AR & \checkmark & \checkmark & \checkmark & \checkmark & \\ 
\rowcolor{lightgreen}
\textbf{\method (ours)} & AR & \checkmark & \checkmark & \checkmark & \checkmark & \checkmark \\ 
\bottomrule
\end{tabular}
}
\label{tab:advantages}
\end{table*}

\subsection{Evaluation Metrics}
\label{app:evaluation_metrics}

For denovo generation: \textbf{Validity.} measures the percentage of generated crystals that pass the structural and chemical validity checks defined by LeMat-GenBench, including physically reasonable geometries and chemically admissible compositions. It reflects the basic feasibility of generated crystal structures. \textbf{Uniqueness.} measures the fraction of generated structures that are structurally distinct from the other generated samples, based on crystal structure matching. A higher value indicates less duplication and greater structural diversity within the generated set. \textbf{Novelty.} measures the fraction of generated structures that cannot be matched to structures in the reference database. It therefore evaluates the ability of a model to explore structures beyond known materials rather than reproducing the training or reference data. \textbf{Formation energy ($E_f$)} measures the energetic preference of a crystal relative to its constituent elemental reference phases. Lower, and particularly more negative, formation energies generally indicate more energetically favorable structures, although they do not alone determine thermodynamic stability against competing compounds. \textbf{Energy above hull ($E_{\mathrm{hull}}$)} measures the energetic distance between a generated material and the convex hull formed by competing phases at the same composition. Lower values indicate greater thermodynamic stability, with structures on the convex hull regarded as stable. \textbf{Relaxation RMSD.} measures the structural displacement between the initially generated crystal and its geometry after energy minimization. A smaller RMSD indicates that the generated structure is already close to a local energy minimum and requires less geometric correction. \textbf{Stable} rate is the percentage of generated structures classified as thermodynamically stable according to their energy above hull. Following LeMat-GenBench, structures with $E_{\mathrm{hull}}\leq 0$ are considered stable. \textbf{Metastable (Meta)} rate measures the percentage of generated structures lying slightly above the convex hull but within the benchmark's metastability window ($0<E_{\mathrm{hull}}\leq0.1$ eV/atom). It captures potentially synthesizable materials that are not strictly ground-state stable. \textbf{S.U.N.} denotes \textbf{Stable, Unique, and Novel} structures. A generated sample is counted only when it is simultaneously thermodynamically stable, structurally distinct from other generated samples, and absent from the reference database, providing a stringent measure of useful material discovery. \textbf{M.S.U.N.} extends S.U.N. to \textbf{Metastable, Unique, and Novel} structures. It measures the fraction of generated candidates that are unique and novel while falling within the predefined metastability window.

Fo symmetry and template metrics: \textbf{Novel Unique Templates.} counts the number of distinct generated Wyckoff templates that do not appear in the reference dataset. Templates are compared independently of elemental identities, such that simply substituting elements into an existing symmetry template does not create a new template. It directly measures the model's ability to explore novel crystallographic arrangements. \textbf{P1 ratio.} reports the percentage of generated structures assigned to space group P1, which contains no nontrivial crystallographic symmetry. Since the corresponding MP-20 reference ratio is $2.23\%$, values close to this reference indicate a realistic balance between low- and high-symmetry structures, whereas a large P1 ratio suggests collapse toward asymmetric structures. \textbf{Space-group $\chi^2$ (SG $\chi^2$)} measures the discrepancy between the generated and reference space-group distributions using a chi-squared statistic. Lower values indicate that the model better reproduces the overall distribution of crystallographic symmetries observed in the reference data. $\boldsymbol{W_{N_{\mathrm{el}}}}$ is the Wasserstein distance between the generated and reference distributions of the number of distinct chemical elements per crystal. It measures whether generated structures exhibit a realistic level of compositional complexity, with lower values indicating better distributional agreement. $\boldsymbol{\mathrm{JSD}_{\mathrm{Wyckoff}}}$ measures the Jensen--Shannon divergence between the generated and reference distributions of occupied Wyckoff positions. Lower values indicate that the model better reproduces the characteristic Wyckoff-site statistics of real crystal structures. \textbf{S.U.N. in symmetry evaluation.} We additionally report S.U.N. in the symmetry analysis to connect crystallographic exploration with material quality. This metric measures how many generated samples are simultaneously stable, unique, and novel rather than merely exhibiting novel symmetry patterns. \textbf{S.S.U.N.} denotes \textbf{Symmetric, Stable, Unique, and Novel} structures. It further requires an S.U.N. candidate to possess nontrivial crystallographic symmetry, i.e., a space group other than P1. S.S.U.N. therefore measures the model's ability to discover genuinely symmetric and energetically viable new materials.

\subsection{More Clarifications on Baseline Comparisons} 

\paragraph{De novo generation.} For de novo generation, we primarily follow the baseline suite provided by LeMat-GenBench to ensure a standardized and reproducible comparison under a common evaluation protocol. We do not include MCFlow because, at the time of our evaluation, we could not identify a reproducible public implementation or released checkpoints that would allow us to generate samples under the same benchmark setting. SGFM is also omitted from the main comparison because it was still available as a preprint rather than an established peer-reviewed publication at the time of evaluation. Moreover, its de novo generation setting is conceptually closer to template-conditioned approaches such as DiffCSP++: SGFM conditions structure generation on Wyckoff positions, and its empirical configuration obtains these symmetry templates by sampling from the distribution observed in the training data rather than generating them jointly with the crystal structure. We acknowledge that additional generic crystal generative models could also be included; however, given the rapidly growing number of diffusion- and flow-based generators with largely similar problem formulations, exhaustively benchmarking every such model is beyond the scope of this work. Our main objective is instead to establish a standardized comparison against representative generic generators while more comprehensively evaluating symmetry-native methods, where \method introduces its primary methodological contribution and demonstrates substantial improvements in symmetry-template exploration and material generation quality.

\paragraph{CSP.} We restrict the symmetry-native CSP comparison in Table~\ref{tab:csp_results} to methods that can natively perform composition- and space-group-conditioned CSP without external template retrieval or separately introduced CSP-specific modules. SymmCD is excluded because its asymmetric-unit formulation jointly generates element types and site symmetries, such that the final stoichiometry depends on the multiplicities of the generated Wyckoff orbits. Consequently, enforcing an arbitrary prescribed composition requires satisfying a global discrete constraint over jointly generated symmetry variables, which is not naturally supported by its diffusion formulation; the authors explicitly identify composition-conditioned CSP as a limitation of the framework. SGEquiDiff likewise does not natively support template-free CSP: its reported CSP pipeline augments the pretrained model with a separate lattice sampler and CSPML-based Wyckoff-template retrieval followed by element substitution, rather than performing CSP through the original generative formulation. The authors further note that native CSP would require modifying the model factorization or introducing composition conditioning during training. SGFM similarly assumes externally supplied Wyckoff information rather than generating the symmetry template from the target composition, with its CSP evaluation relying on ground-truth or CSPML-retrieved Wyckoff positions. We therefore compare against CrystalFormer and WyFormer, whose released implementations provide native composition- and space-group-conditioned CSP pipelines with explicit symmetry-template generation.

\subsection{Reproducibility Settings} 
\label{app:reproducibility_settings}

\paragraph{Data and model configurations.}
We use three task-specific models for MP-20 de novo generation, MP-20 CSP, and MPTS-52 CSP.
Symmetry preprocessing uses a positional tolerance of $0.01$~\AA{} and an angular tolerance of $5^\circ$.
Structures that cannot be represented with a verified symmetry reconstruction are retained through a P1 representation.
Table~\ref{tab:reproducibility_configurations} summarizes the three configurations.
The two MP-20 models share architectural dimensions but use separately trained base weights.

\begin{table}[t]
\centering
\caption{Configurations of the three models used for the reported results.}
\label{tab:reproducibility_configurations}
\small
\setlength{\tabcolsep}{6pt}
\begin{tabular}{@{}lccc@{}}
\toprule
Setting & De novo & CSP: MP-20 & CSP: MPTS-52 \\
\midrule
Parameters (M) & 32.804 & 33.968 & 91.807 \\
Hidden width & 384 & 384 & 512 \\
Template/coordinate layers & 5/5 & 5/5 & 8/8 \\
Attention heads & 6 & 6 & 8 \\
Feed-forward width & 1,536 & 1,536 & 2,048 \\
Lattice-encoder layers & 4 & 4 & 6 \\
Base-training updates & 20,000 & 20,000 & 40,000 \\
CSP loss weight in base training & 0.25 & 0.25 & 1.0 \\
CSP adapter updates & --- & 4,000 & 3,000 \\
Orbit-count-head updates & --- & 4,000 & 4,000 \\
Coordinate beam/expansion width & 4/56 & 16/32 & 16/32 \\
Coordinate prefix-evaluation budget & 2,049 & 512 & 256 \\
\bottomrule
\end{tabular}
\end{table}

\paragraph{Base training.}
We use AdamW with learning rate $3\times10^{-4}$, $(\beta_1,\beta_2)=(0.9,0.95)$, weight decay $0.01$, and gradient-norm clipping at $1.0$.
The learning-rate schedule combines a 500-step linear warm-up with cosine decay to $3\times10^{-5}$.
Training uses BF16 mixed precision and an effective batch size of 128.

\paragraph{De novo generation.}
It conditions generation on an empirical joint prior over the space group, total atom count, number of inequivalent sites, and number of distinct elements, estimated from the training split.
This configuration fixes the template length through the sampled prior.
We generate 2,500 fixed attempts with template and lattice temperatures of $1.0$, and coordinate temperature $0.7$.
Coordinate decoding uses simulation-guided beam search with the budget in Table~\ref{tab:reproducibility_configurations} and greedy legal completions of partial paths.

\paragraph{Relaxation and de novo evaluation.}
The raw and pre-relaxed results use the same 2,500 attempts, with unsuccessful attempts retained in the denominator.
The reported energy and stability statistics first average the three MLIP energies for each structure.
Stable structures satisfy $E_{\mathrm{hull}}\leq0$, and the metastable interval is $0<E_{\mathrm{hull}}\leq0.1$~eV/atom.
Energy and relaxation-RMSD statistics are computed over available valid structures, whereas all percentages retain the original attempt denominator.

\paragraph{CSP adaptation.}
We freeze their base weights and train rank-32 CSP adapters using a loss that doubles the template cross-entropy over ambiguous, composition-feasible choices while retaining the lattice and coordinate losses.
Adapter training uses AdamW with learning rate $3\times10^{-4}$, 100-step warm-up, $(\beta_1,\beta_2)=(0.9,0.95)$, weight decay $0.01$, gradient clipping at $1.0$, and effective batch size 128.
An auxiliary head predicts the number $M$ of inequivalent sites from the template context.
This head has hidden width 128 and is trained for 4,000 updates with batch size 256, learning rate $10^{-3}$, AdamW coefficients $(0.9,0.999)$, and weight decay $0.01$.

We will open-source an implementation of \method, together with model configurations, inference scripts, evaluation code, and generation results, in \href{https://crssbl.github.io/Project-CrystalAR/\#results}
{\nolinkurl{https://crssbl.github.io/Project-CrystalAR/\#results}}

\subsection{Ablation Studies of de novo}
\label{sec:ablation_denovo}

We group the de novo ablations into count and search policy, representation and module components, and fixed-skeleton chemistry interventions. All rates below use 2,500 attempted structures. Validity, uniqueness, and structural novelty are computed on the submitted structures. Stability and S.U.N. use the mean $E_{\mathrm{hull}}$ from Orb, MACE, and UMA; failed attempts remain in the denominator.

\paragraph{Count prior and constrained search.}
Table~\ref{tab:ablation_prior_search_denovo} compares the empirical count prior used by \method with a learned stopping policy. The empirical prior raises the SGBS stable rate from 3.52\% to 6.00\% and S.U.N. from 4 to 7 structures, while the learned stopping policy gives higher structural novelty (48.92\% versus 42.76\%). SGBS increases the \method valid rate from 64.40\% to 91.44\% and structural novelty from 18.76\% to 42.76\%; S.U.N. increases from 5 to 7 structures. Orb pre-relaxation applied to the same \method SGBS cohort raises the stable rate to 13.68\% and S.U.N. to 22 structures (0.88\%).

\begin{table}[t]
\centering
\caption{\textbf{Count prior and search ablation for de novo generation.} All entries are percentages over 2,500 attempts; S.U.N. counts are shown in parentheses.}
\label{tab:ablation_prior_search_denovo}
\small
\setlength{\tabcolsep}{4pt}
\begin{tabular}{@{}llrrrrr@{}}
\toprule
Count policy & Search & Valid & Unique & Novel & Stable & S.U.N. \\
\midrule
Empirical prior (\method) & Sampling & 64.40 & 62.24 & 18.76 & 5.44 & 0.20 (5) \\
Empirical prior (\method) & SGBS & 91.44 & 89.08 & 42.76 & 6.00 & 0.28 (7) \\
Learned STOP & Sampling & 59.96 & 58.72 & 21.72 & 3.48 & 0.16 (4) \\
Learned STOP & SGBS & 90.12 & 88.64 & 48.92 & 3.52 & 0.16 (4) \\
Composition prefix & SGBS & 91.44 & 89.52 & 41.32 & 5.40 & 0.20 (5) \\
\bottomrule
\end{tabular}
\end{table}

\paragraph{Representation and key modules.}
Table~\ref{tab:ablation_components_denovo} reports representative interventions. Full-cell autoregressive generation directly models every atom and reaches 49.12\% validity, 48.88\% uniqueness, 35.04\% novelty, 0.72\% stability, and no S.U.N. structure. The Wyckoff-letter coordinate representation retains 89.80\% validity and reaches 51.36\% novelty, while the flat and continuous coordinate heads reduce validity to 68.16\% and 41.64\%, respectively. Removing the template-state bridge produces 96.28\% validity and 8.68\% stability together with 19.32\% novelty and one S.U.N. structure. This separation between geometric validity and stable novelty highlights the role of state coupling in the generation process. Removing periodic encoding gives 0 S.U.N. structures, and removing the condition query gives 3.

\begin{table}[t]
\centering
\caption{\textbf{Representative representation and module ablations.} Rates are percentages over 2,500 attempts; S.U.N. counts are shown in parentheses.}
\label{tab:ablation_components_denovo}
\small
\setlength{\tabcolsep}{4pt}
\begin{tabular}{@{}lrrrrr@{}}
\toprule
Configuration & Valid & Unique & Novel & Stable & S.U.N. \\
\midrule
Learned STOP reference & 90.12 & 88.64 & 48.92 & 3.52 & 0.16 (4) \\
Full-cell AR & 49.12 & 48.88 & 35.04 & 0.72 & 0.00 (0) \\
Wyckoff-letter coordinates & 89.80 & 88.08 & 51.36 & 4.16 & 0.08 (2) \\
Flat 512-bin coordinates & 68.16 & 66.48 & 29.12 & 4.24 & 0.00 (0) \\
Continuous coordinates & 41.64 & 40.84 & 17.12 & 1.72 & 0.08 (2) \\
No periodic encoding & 89.96 & 88.04 & 50.00 & 4.12 & 0.00 (0) \\
No condition query & 90.52 & 89.08 & 49.72 & 3.72 & 0.12 (3) \\
No template-state bridge & 96.28 & 90.96 & 19.32 & 8.68 & 0.04 (1) \\
\bottomrule
\end{tabular}
\end{table}

\paragraph{Fixed-skeleton chemistry trade-off.}
We preserve each generated space-group and Wyckoff skeleton and vary the element sampling temperature. Increasing the temperature from 1.0 to 2.5 raises novelty from 41.84\% to 57.92\%, including an increase in structures with formulas absent from the reference set from 489 to 802. Validity falls from 87.44\% to 78.72\%, and stability falls from 5.56\% to 1.60\%. The corresponding S.U.N. count changes from 1 to 3. Removing the clearance cap at the higher temperature raises novelty to 58.44\% while reducing stability to 0.96\%; S.U.N. remains at 3. These results describe a chemical exploration--stability trade-off at fixed symmetry support.

\begin{table}[t]
\centering
\caption{\textbf{Fixed-skeleton chemistry interventions.} All arms use the same generated symmetry skeleton and 2,500 attempts; rates are percentages and S.U.N. counts are shown in parentheses.}
\label{tab:ablation_fixed_skeleton_denovo}
\small
\setlength{\tabcolsep}{4pt}
\begin{tabular}{@{}lrrrrr@{}}
\toprule
Chemistry intervention & Valid & Unique & Novel & Stable & S.U.N. \\
\midrule
Original elements & 88.36 & 86.00 & 39.68 & 5.92 & 0.28 (7) \\
Element redraw, $T_E=1.0$ & 87.44 & 85.24 & 41.84 & 5.56 & 0.04 (1) \\
Element redraw, $T_E=2.5$ & 78.72 & 78.60 & 57.92 & 1.60 & 0.12 (3) \\
$T_E=2.5$ with uncapped clearance & 78.72 & 78.60 & 58.44 & 0.96 & 0.12 (3) \\
\bottomrule
\end{tabular}
\end{table}

\begin{figure}[t]
\centering
\includegraphics[width=\textwidth]{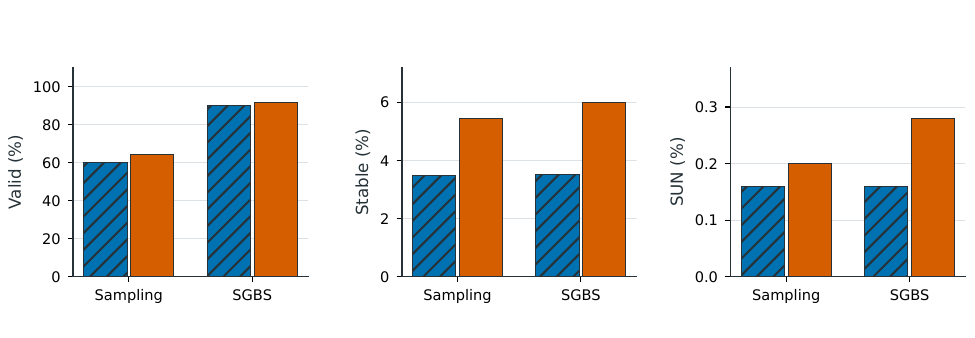}
\caption{\textbf{Count prior and constrained search.} The panels show validity, stability, and S.U.N. for sampling and SGBS. CrystalAR uses the empirical count prior; the learned-STOP reference uses a learned stopping policy. All values are computed over 2,500 attempts, and the exact percentages and counts appear in Table~\ref{tab:ablation_prior_search_denovo}.}
\label{fig:ablation_prior_search_denovo}
\end{figure}

\begin{figure}[t]
\centering
\includegraphics[width=\textwidth]{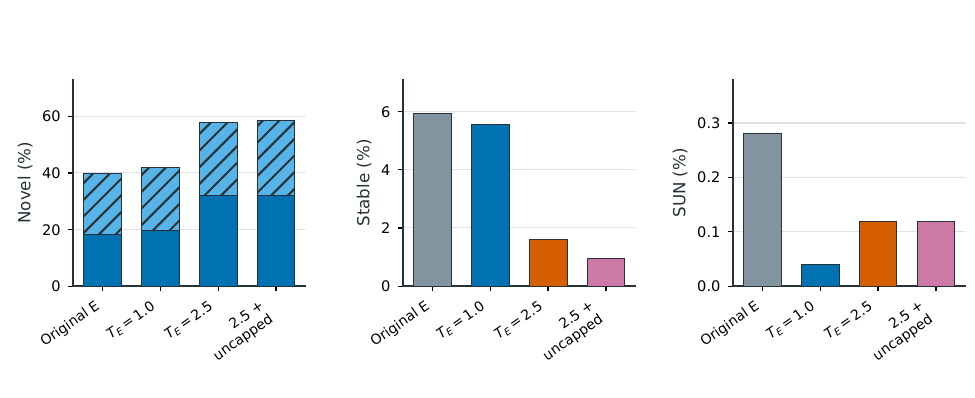}
\caption{\textbf{Novelty--stability trade-off under fixed symmetry support.} The panels show novelty, stability, and S.U.N. for the four chemistry interventions in Table~\ref{tab:ablation_fixed_skeleton_denovo}. The element sampling temperature changes chemical exploration while the generated symmetry skeleton is held fixed.}
\label{fig:ablation_novelty_tradeoff_denovo}
\end{figure}

\subsection{Ablation Studies of CSP}

\begin{figure}[t]
\centering
\includegraphics[width=\textwidth]{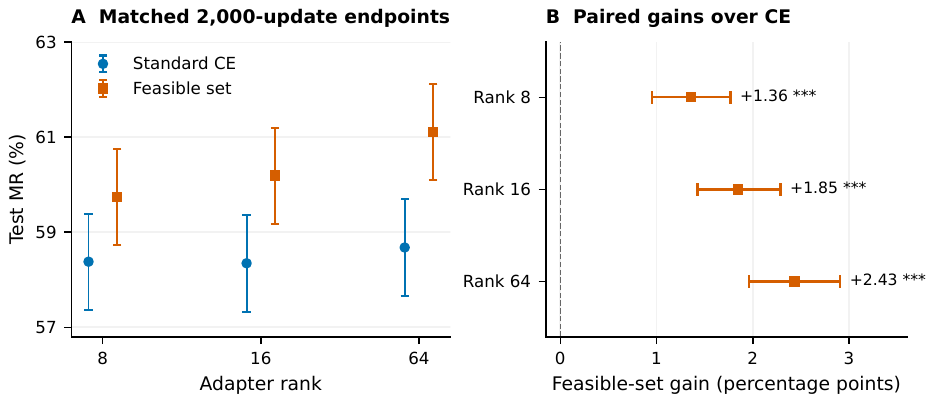}
\caption{\textbf{Feasible-set supervision improves CSP across adapter capacities.} Both panels use 2,000 adaptation updates and 9,046 structures. (A) MR for standard CE and feasible-set supervision, with Wilson 95\% confidence intervals. (B) Paired MR gains, with bootstrap 95\% confidence intervals; $^{***}$ indicates $p<0.001$ under exact McNemar tests with Holm correction.}
\label{fig:ablation_csp_training}
\end{figure}

\paragraph{CSP training objective and adapter capacity.}
We examine how supervision, adaptation capacity, and decoding affect composition- and space-group-conditioned CSP on MP-20.
The training comparisons in Table~\ref{tab:ablation_csp_training} use autoregressive template ranking and a fixed coordinate-search budget.
Both adapter objectives retain the lattice and coordinate losses and double the template-loss weight.
Standard cross-entropy (CE) uses the full output vocabulary.
The feasible-set objective normalizes over composition-feasible choices and supervises ambiguous element, site-symmetry, and coordinate-form decisions.
At 2,000 updates, this objective improves MR over CE by 1.36, 1.85, and 2.43 percentage points at ranks 8, 16, and 64, respectively (Figure~\ref{fig:ablation_csp_training}).
The gains correspond to 123, 167, and 220 additional recovered structures, supporting the benefit of aligning supervision with the feasible choices encountered during generation.

\begin{table}[t]
\centering
\caption{\textbf{CSP adaptation with fixed single-hypothesis decoding on MP-20.} All configurations use autoregressive template scores and the same coordinate-search settings. Updates denote the adaptation budget, and rank denotes adapter capacity. RMSE is averaged over matched structures. 
}
\label{tab:ablation_csp_training}
\footnotesize
\setlength{\tabcolsep}{6pt}
\renewcommand{\arraystretch}{1.10}
\begin{adjustbox}{max width=\textwidth}
\begin{tabular}{@{}lrrrr@{}}
\toprule
Adaptation & Rank & Updates & MR (\%) & RMSE \\
\midrule
\multicolumn{5}{l}{\textit{Reference}} \\
Unadapted model & --- & 0 & 57.95 & 0.05699 \\
\midrule
\multicolumn{5}{l}{\textit{Standard template CE}} \\
CE adapter & 8 & 2,000 & 58.38 & 0.05863 \\
CE adapter & 16 & 2,000 & 58.35 & 0.05797 \\
CE adapter & 32 & 2,000 & 58.42 & 0.05733 \\
CE adapter & 64 & 2,000 & 58.68 & 0.05830 \\
CE adapter & 32 & 4,000 & 58.57 & 0.05750 \\
\midrule
\multicolumn{5}{l}{\textit{Feasible-set supervision}} \\
Feasible-set adapter & 8 & 2,000 & 59.74 & 0.05751 \\
Feasible-set adapter & 16 & 2,000 & 60.19 & 0.05767 \\
Feasible-set adapter & 64 & 2,000 & 61.11 & 0.05851 \\
Feasible-set adapter & 32 & 4,000 & 61.35 & 0.05586 \\
\midrule
\multicolumn{5}{l}{\textit{Joint optimization of de novo and CSP objectives}} \\
Full-model joint & --- & 2,000 & 56.89 & 0.05938 \\
Full-model joint & --- & 4,000 & 57.13 & 0.06013 \\
\bottomrule
\end{tabular}
\end{adjustbox}
\end{table}

\paragraph{Adaptation budget and parameter sharing.}
Doubling the rank-32 CE budget from 2,000 to 4,000 updates increases MR from 58.42\% to 58.57\%, a gain of 0.14 percentage points with a paired 95\% interval of $[-0.24,0.53]$.
The rank-32 feasible-set configuration reaches 61.35\% MR and 0.05586 RMSE at 4,000 updates.
Full-model joint optimization of de novo and CSP objectives yields 56.89\% and 57.13\% MR at the two budgets, compared with 57.95\% for the unadapted model.
Under these optimization settings, CSP-only feasible-set adaptation provides the strongest structure recovery while keeping the shared model parameters fixed.
The joint and adapter configurations use their respective loss weighting and learning-rate settings; these results characterize the complete adaptation strategies.

\begin{figure}[t]
\centering
\includegraphics[width=\textwidth]{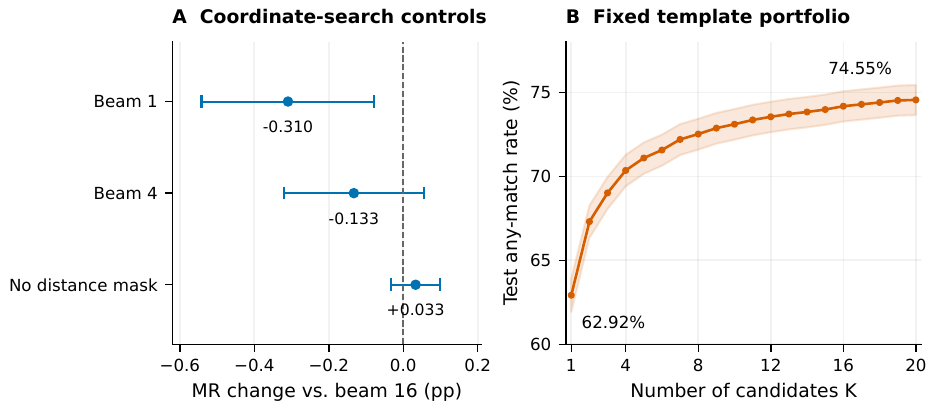}
\caption{\textbf{Effect of coordinate search and hypothesis count on CSP.} (A) MR changes relative to coordinate beam 16 with the distance mask, with paired bootstrap 95\% confidence intervals. (B) MR as the number of generated hypotheses increases, with a Wilson 95\% confidence band over 9,046 structures.}
\label{fig:ablation_csp_search}
\end{figure}

\paragraph{Coordinate-search controls.}
We vary coordinate decoding while fixing the model parameters and template ranking, including the auxiliary orbit-count score (Table~\ref{tab:ablation_csp_inference}).
Reducing the coordinate beam width from 16 to 4 or 1 changes MR from 62.92\% to 62.79\% or 62.61\%, respectively (Figure~\ref{fig:ablation_csp_search}A).
All beam widths retain simulation rollouts and the same expansion and search budgets.
Removing the distance mask changes MR by $+0.03$ percentage points, with a paired 95\% interval of $[-0.03,0.10]$, and decreases structural validity from 99.44\% to 98.91\%.
These results indicate modest sensitivity to coordinate beam width and a contribution of distance constraints to geometric validity.

\begin{table}[t]
\centering
\caption{\textbf{Coordinate-search ablations for MP-20 CSP.} Model parameters, template ranking, and the remaining decoding settings are fixed. Structural validity measures geometric validity and composition consistency. MR additionally applies the composition-validity criteria used in the CSP evaluation.}
\label{tab:ablation_csp_inference}
\footnotesize
\setlength{\tabcolsep}{6pt}
\renewcommand{\arraystretch}{1.10}
\begin{adjustbox}{max width=\textwidth}
\begin{tabular}{@{}lrrr@{}}
\toprule
Coordinate decoding & MR (\%) & RMSE & Structural valid (\%) \\
\midrule
Beam 16 (default) & 62.92 & 0.05847 & 99.44 \\
Beam 4 & 62.79 & 0.05864 & 99.56 \\
Beam 1 & 62.61 & 0.05850 & 99.60 \\
Beam 16, distance mask removed & 62.96 & 0.05860 & 98.91 \\
\bottomrule
\end{tabular}
\end{adjustbox}
\end{table}

\paragraph{Number of generated hypotheses.}
Increasing the candidate budget from 1 to 4, 8, and 20 hypotheses raises MR from 62.92\% to 70.35\%, 72.52\%, and 74.55\%, respectively (Figure~\ref{fig:ablation_csp_search}B).
The 20-hypothesis setting recovers 6,744 structures, adding 1,052 recoveries over single-hypothesis generation.
Multi-hypothesis MR measures the fraction of inputs with at least one accepted match among these candidates.
The curve shows diminishing returns as the candidate budget increases, with eight hypotheses reaching 72.52\% MR.

\begin{figure}[t]
\centering
\includegraphics[width=\textwidth]{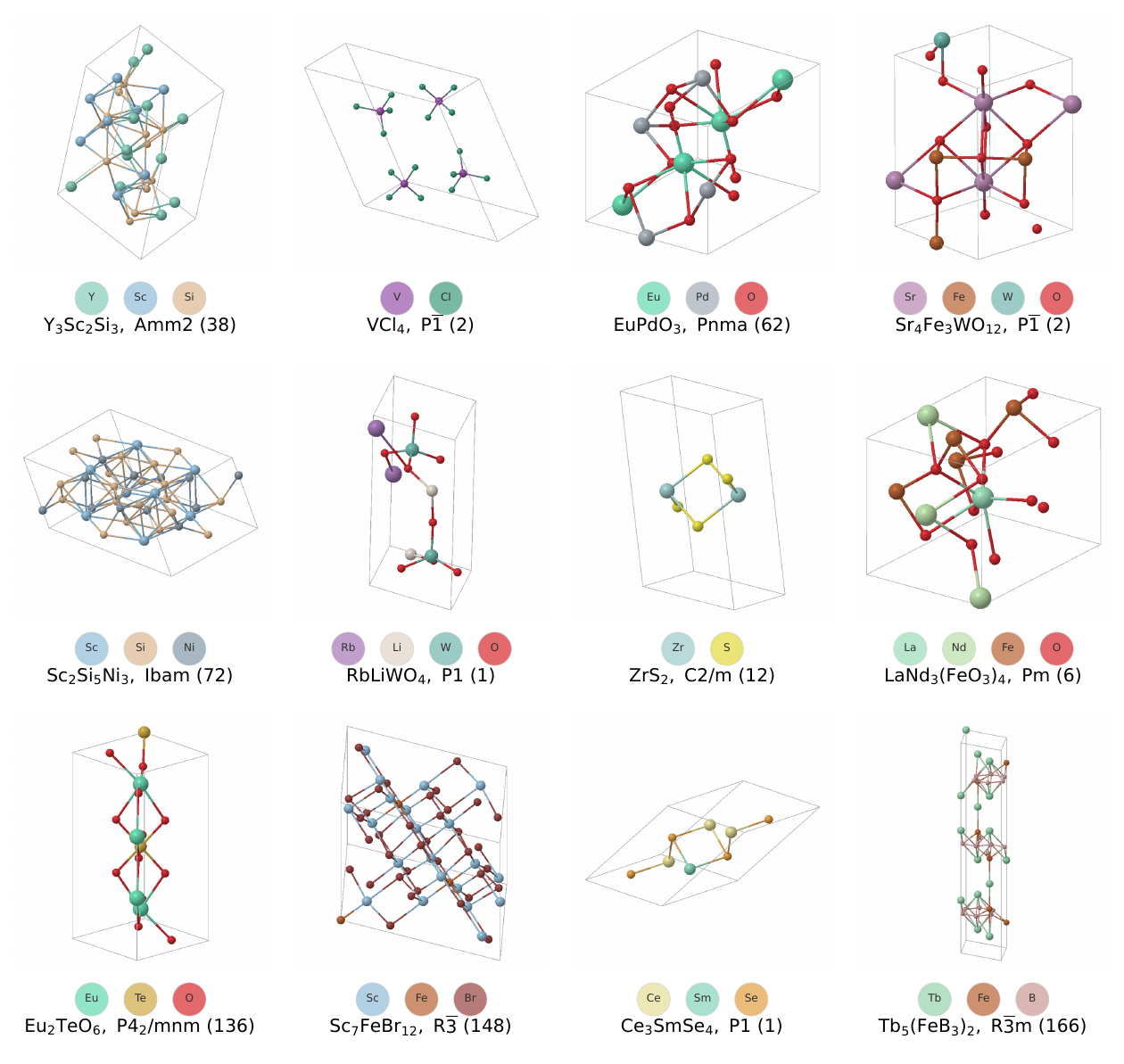}
\caption{\textbf{Visualization.} More S.U.N structures generated by \method.}
\label{fig:more_sun_vis}
\end{figure}

\subsection{Further Clarifications on Template-free Crystal Structure Prediction} 

We clarify the distinction between conventional composition-conditioned CSP and template-free CSP under symmetry-native generation. Generic generative models, such as DiffCSP and FlowMM, predict crystal structures conditioned only on chemical composition, whereas symmetry-native models explicitly incorporate crystallographic symmetry into their generative formulations and are typically evaluated with a specified space group \(G\). Since specifying \(G\) provides additional crystallographic information, the two settings are not strictly equivalent in terms of conditioning. However, the purpose of this comparison is not to establish information-equivalent performance between generic and symmetry-native models, but rather to investigate how much crystallographic prior information can be relaxed for symmetry-native CSP.

In particular, previous symmetry-native approaches, such as DiffCSP++ and SGEquiDiff, rely on template retrieval to provide both the space group \(G\) and the Wyckoff template \(W\). The latter specifies the symmetry-inequivalent atomic sites, their Wyckoff multiplicities, and associated positional constraints, thereby substantially restricting the space of possible crystal structures. In contrast, specifying only \(G\) defines the underlying symmetry constraints without determining how atomic species are distributed across Wyckoff positions. Consequently, inferring \(W\) from chemical composition and \(G\) remains a nontrivial combinatorial problem involving stoichiometric compatibility, site multiplicities, and crystallographic constraints.

Importantly, relaxing the requirement for a predefined Wyckoff template is not merely a reduction in prior information, but also has practical implications for crystal discovery. For previously unexplored compositions, suitable Wyckoff templates may be unavailable in existing crystal databases, limiting the applicability of retrieval-based approaches. Moreover, even for a given composition and space group, multiple symmetry-compatible Wyckoff configurations may exist, potentially corresponding to distinct crystal structures or polymorphs. Restricting CSP to retrieved templates may therefore bias structural exploration toward previously observed crystallographic configurations and overlook alternative arrangements. By explicitly modeling the generation of \(W\), \method removes this dependency and enables the exploration of diverse symmetry-compatible templates jointly with their corresponding crystal geometries. This represents a meaningful step toward more flexible symmetry-native CSP, extending its potential applicability to previously unexplored material systems and polymorph discovery while preserving crystallographic symmetry throughout generation.

\subsection{Limitations}

Despite its strong performance, \method has two main limitations. First, we observe a noticeable stability degradation when generating crystals with highly novel Wyckoff templates. This reflects a broader challenge shared by symmetry-native generative models that explicitly explore novel crystallographic templates: moving beyond templates well represented in the training distribution substantially enlarges the structural search space, making it more difficult to identify energetically favorable realizations. Improving the discovery of stable structures under novel symmetry templates therefore remains an important direction for future work. Second, the autoregressive formulation of \method naturally supports beam search, providing a flexible mechanism for exploring multiple high-probability structural hypotheses that is less straightforward for diffusion-based generation. However, this advantage introduces a quality--efficiency trade-off: increasing the beam size generally improves search quality while proportionally increasing inference cost. Future work could reduce this dependence on exhaustive search through post-training techniques, such as reinforcement learning or preference-based optimization, to better concentrate probability mass on high-quality candidates and achieve more efficient decoding with smaller search budgets. 

\end{document}